# Quantum-Classical Physics-Informed Neural Networks for Solving Reservoir Seepage Equations

Xiang Rao[1,2,3,4,6]*(饶翔), Yina Liu[5]*(刘怡娜), Yuxuan Shen[4](申毓萱)
[1]School of Petroleum Engineering, Yangtze University, Wuhan 430100, China
[2]School of Computer Science, Yangtze University, Jingzhou 434023, China
[3]State Key Laboratory of Low Carbon Catalysis and Carbon Dioxide Utilization (Yangtze University), Wuhan 430100, China
[4]College of Future Technology, Yangtze University, Wuhan 430100, China
[5]King Abdullah University of Science and Technology, Physical Science and Engineering Division, 23955-6900, Saudi Arabia
[6]Western Research Institute, Yangtze University, Karamay 834000, China
*Corresponding author: Prof. Xiang Rao (raoxiang0103@163.com, raoxiang@yangtzeu.edu.cn), Miss. Yina Liu (yina.liu@kaust.edu.sa)

**Abstract**: Solving reservoir seepage partial differential equations (PDEs) is critical for hydrocarbon development optimization and production forecasting. Conventional numerical solvers suffer from mesh errors and high computational cost. Standard Physics-Informed Neural Networks (PINNs) are limited by inefficient parameters. They also face weak high-dimensional representation and poor nonlinear fitting ability. We adapt the Discrete Variable (DV)-Circuit Quantum-Classical Physics-Informed Neural Network (QCPINN) to tackle these drawbacks. This work applies the method to four representative reservoir seepage models. The models cover heterogeneous single-phase pressure diffusion. They also include nonlinear Buckley-Leverett two-phase waterflooding. Another case is adsorption-coupled convection-diffusion compositional flow. The last one refers to fully coupled pressure-saturation two-phase flow with exponential permeability heterogeneity. The QCPINN combines classical preprocessing and postprocessing networks with a DV quantum core. It leverages quantum entanglement and superposition to boost high-dimensional feature mapping. Physical constraints are embedded within the framework to guarantee physically consistent solutions. We evaluate three quantum circuit topologies including Cascade, Cross-mesh and Alternate. Numerical tests prove QCPINNs outperform classical PINNs in prediction accuracy, and the circuit topologies show case-specific performance with clear applicable ranges. This work verifies QCPINN's practicability for reservoir engineering. It bridges quantum computing research with oilfield industrial applications.

## 1 Introduction

Solving partial differential equations (PDEs) is a core problem in the fields of science and engineering. The accurate solution of PDEs for reservoir seepage directly determines the reliability of oil and gas field development plan optimization and production performance prediction. Traditional numerical methods, such as the finite volume method (Rao et al., 2024), finite element method (Jackson et al., 2015), meshless method (Rao et al., 2022), and others (Wang et al., 2023; Rao et al., 2025a), rely on mesh or point cloud discretization. When dealing with strong reservoir heterogeneity, strong flow nonlinearity, and convection-dominated problems, these methods are prone to mesh-dependent errors or numerical dispersion. Furthermore, the computational cost increases exponentially with the complexity of the system, making it difficult to meet the demand for efficient simulation of complex oil and gas reservoirs.

Physics-Informed Neural Networks (PINNs) embed physical laws into the neural network architecture (Raissi et al., 2019) and achieve PDE solving through a combination of data-driven and physics-constrained paradigms. They can ensure the physical consistency of solutions without requiring large amounts of labeled data, providing a new approach for reservoir forward simulation and inverse problem solving (Almajid & Abu-Al-Saud, 2022; Hanna et al., 2022). Recently, Physics-Informed Kolmogorov-Arnold Networks (PIKAN) have also been developed for reservoir flow simulation (Rao et al., 2025b; Kalesh et al., 2025). For instance, Rao et al. (2025b) applied PIKAN to single-phase seepage simulation of heterogeneous permeability based on a mixed pressure-velocity formulation and found that PIKAN outperformed PINNs in simulation performance. However, classical PINNs and PIKANs face three core limitations when addressing complex reservoir flow scenarios: low parameter efficiency, insufficient high-dimensional expression capability, and restricted fitting of strong nonlinearity. First, to capture the heterogeneity of permeability or the high-dimensional characteristics of governing equations, it is necessary to design deep fully connected networks or increase the number of neurons, leading to a surge in the scale of trainable parameters (usually ranging from thousands to tens of thousands). This not only increases the memory and time costs of training but also easily causes overfitting and unstable convergence. Second, when simulating the steep shock front of convection-dominated equations, the introduction of artificial viscosity terms (Fuks & Tchelepi, 2020) and the gradient propagation characteristics of classical activation functions (e.g., Tanh, ReLU) tend to result in gradient vanishing or local oscillations, making it difficult to balance accuracy and convergence speed. These limitations prevent classical PINNs from achieving both efficiency and accuracy in complex reservoir simulations, becoming a key bottleneck restricting their industrial application.

As a revolutionary paradigm to break through the bottlenecks of classical computing, quantum computing can trace its development back to the 1980s. Feynman (1982) proposed the core idea of quantum system simulation, and Deutsch (1985) confirmed the universal Turing machine property of quantum computers, laying a theoretical foundation for quantum computing. Subsequent landmark achievements, such as the Deutsch-Jozsa algorithm (1992), Shor's large integer factorization algorithm (1994), Grover's database search algorithm (1996), and the Harrow-Hassidim-Lloyd (HHL) linear system solving algorithm (Harrow et al., 2009), have theoretically verified the exponential acceleration advantages of quantum computing in specific problems. At the experimental level, Cirac and Zoller (1995) proposed the ion trap qubit scheme, Monroe et al. (1995) implemented the first CNOT quantum logic gate, and Chuang et al. (1998) verified the Deutsch-Jozsa algorithm based on nuclear magnetic resonance technology, promoting quantum computing from theory to experiment.

Currently, quantum computing is in the Noisy Intermediate-Scale Quantum (NISQ) era (Bharti et al., 2022), and variational quantum algorithms tailored for NISQ hardware have become a research focus (Cerezo et al., 2021a), including the Variational Quantum Eigensolver (VQE) (Peruzzo et al., 2014; Tilly et al., 2022), Variational Quantum Linear Solver (VQLS) (Bravo-Prieto et al., 2023), Variational Quantum Classifier (VQC) (Havlíček et al., 2019; Jäger and Krems., 2023), and Variational Quantum Phase Estimation (VQPE) (Liu et al., 2024). These algorithms generate quantum states through Parameterized Quantum Circuits (PQC) and iteratively optimize parameters using classical optimizers. While avoiding noise accumulation in deep quantum circuits, they achieve approximate solutions to complex problems. In the field of PDE solving, VQLS has been attempted for solving Poisson equations (Ali and Kabel, 2023; Govindugari and Wong, 2024; Ayoub and Baeder, 2025), heat conduction equations (Liu et al., 2022), reservoir seepage equations (Rao., 2024a; Rao., 2024b), and others. By decomposing the coefficient matrix of the linear system (obtained from PDE numerical discretization) into a superposition of Pauli bases, VQLS computes the cost function using quantum state preparation and Hadamard testing to obtain the solution of the linear system. Therefore, the application of VQLS to PDE computation requires prior numerical discretization of the PDE, rather than obtaining the PDE solution purely through quantum computing. Quantum neural networks (QNNs) based on variational quantum algorithms have also been used as pure machine learning methods (Tüysüz et al., 2021; Rao et al., 2024b); for example, Rao et al. (2024b) applied QNNs to predict carbon dioxide ($CO_2$) sequestration in saline aquifers. However, in general, due to the lack of physical information and the more challenging nonlinear processing compared to classical networks, the prediction performance of pure QNNs is relatively limited.

A natural idea is to develop pure Quantum Physics-Informed Neural Networks (QPINNs). For example, Trahan et al. (2024) applied QPINNs to spring-mass systems and Poisson equations, while Berger et al. (2025) applied QPINNs to Poisson, Burger's, and Navier-Stokes equations. Xiao et al. (2024) used QPINNs to solve forward and inverse problems of certain PDEs. Theoretically, however, pure quantum neural networks require more layers than hybrid quantum-classical neural networks, which significantly increases the hardware threshold for physical implementation of the network and makes it more susceptible to noise interference. Moreover, hybrid quantum-classical neural networks can leverage the advantage that classical networks achieve nonlinear complexity at a much lower cost than quantum networks, thereby enhancing the nonlinear fitting capability of the neural network.

Thus, hybrid Quantum-Classical Physics-Informed Neural Networks (QCPINNs) have been further developed. For instance, Sedykh et al. (2024) proposed a QCPINN for simulating laminar fluid flow in 3D Y-shaped mixers, and Dehaghani et al. (2024) applied QCPINNs to solve quantum optimal control problems. Fernandez et al. (2025) integrated a quantum layer into a classical PINN to form a QCPINN and conducted preliminary tests on its computational performance for wave equations. Trahan et al. (2024) and Leong et al. (2025) applied QCPINNs to 1D Burger's equations and 2D inviscid compressible flows, respectively. Farea et al. (2025) applied QCPINNs to Helmholtz equations, wave equations, Klein-Gordon equations, and other PDEs. Through numerical experiments, both Trahan et al. (2024) and Farea et al. (2025) found that compared with classical PINNs, QCPINNs can construct high-dimensional feature spaces using the properties of quantum superposition and entanglement. This enables PDE solving with far fewer parameters than classical PINNs, significantly improving parameter efficiency. QCPINNs provide a breakthrough solution to the core limitations of classical PINNs: through a hybrid architecture of classical preprocessing, quantum core computing, and classical postprocessing, they achieve deep integration of quantum advantages and physical constraints.

As a newly developed method in the past two years, QCPINNs have so far been applied primarily to fundamental physical PDEs and have not been extended to the field of reservoir engineering. Therefore, this study for the first time applies QCPINN technology to reservoir engineering, systematically simulating four typical reservoir seepage models: the pressure diffusion equation for single-phase reservoir flow with heterogeneous permeability, the Buckley-Leverett (BL) equation for simplified two-phase waterflooding, and the convection-diffusion equation for component concentration considering adsorption in compositional flow, and the fully coupled pressure-saturation two-phase seepage equation for heterogeneous reservoirs with exponential permeability distribution. It accurately captures core features such as permeability heterogeneity, flow nonlinearity, and convection dominance, providing an early test case for the future development of reservoir numerical simulators or machine learning surrogate models based on quantum computing. This work promotes the translation of quantum computing from basic physics research to industrial application in oil and gas engineering.

## 2 Typical Partial Differential Equations in Reservoir Numerical Simulation

This section briefly presents the corresponding typical partial differential equations: the pressure diffusion equation for heterogeneous permeability in single-phase flow, the nonlinear Buckley-Leverett (BL) equation for two-phase flow, and the convection-diffusion equation for component concentration considering adsorption effects.

### 2.1 Pressure Diffusion Equation with Heterogeneous Permeability in Single-Phase Flow

The core of single-phase flow (e.g., pure oil-phase flow in the early stage of reservoir development) lies in the diffusion process of the pressure field in a medium with heterogeneous permeability. Its governing equation is derived based on Darcy's law and the law of mass conservation. For a two-dimensional (2D) reservoir domain $\mathbf{x}=(x,y)$ and time domain $t\in[0,T]$, the general form of the pressure diffusion equation for heterogeneous single-phase flow is:

$$\frac{\partial}{\partial t}\big(\phi(\mathbf{x})\rho\big)=\nabla\cdot\left(\frac{k(\mathbf{x})}{\mu}\rho\nabla p\right)+q_s(\mathbf{x},t), \tag{1}$$

where $p(\mathbf{x},t)$ denotes reservoir pressure, the unknown function to be solved; $\phi(\mathbf{x})$ is porosity; $k(\mathbf{x})$ represents the absolute permeability of the rock; $\rho$ is fluid density, assumed constant under weakly compressible conditions; $\mu$ denotes fluid viscosity; $q_s(\mathbf{x},t)$ is the source/sink term (negative for production wells where fluid flows out, positive for injection wells where fluid flows in). Under steady-state and weakly compressible fluid conditions, Eq. (1) can be further simplified to an elliptic equation for pressure:

$$\nabla\cdot\left(\frac{k(\mathbf{x})}{\mu}\nabla p\right)+\frac{q_s(\mathbf{x})}{\rho}=0, \tag{2}$$

Boundary conditions include Dirichlet boundaries (constant pressure boundaries, such as edge water or bottom water boundaries) and Neumann boundaries (constant flux boundaries, such as closed faults or impermeable boundaries), which satisfy $p(\mathbf{x})=p_\mathrm{b}$, $\left(\mathbf{x}\in\Gamma_D, t\in[0,T]\right)$ and $\nabla p(\mathbf{x})\cdot\mathbf{n}=0$, $\left(\mathbf{x}\in\Gamma_N, t\in[0,T]\right)$, respectively. Here, $\Gamma_D\subset\partial\Omega$ and $\Gamma_N\subset\partial\Omega$ are the constant pressure and constant flux boundaries, $p_b$ is the boundary pressure, and $\mathbf{n}$ is the boundary normal vector. The spatial heterogeneity of permeability $k(\mathbf{x})$ is the key factor leading to the nonlinearity of pressure distribution.

### 2.2 Nonlinear Buckley-Leverett (BL) Equation in Two-Phase Flow

The waterflooding process is a typical two-phase flow, and the BL equation describes the nonlinear convection process of saturation through simplified mass conservation relationships, requiring special handling of the steep gradient at the shock front. Under the one-dimensional (1D) flow assumption (along the main streamline direction $x$), ignoring gravity and capillary forces, the conservative form of the BL equation is:

$$\frac{\partial S_w}{\partial t}+v\frac{\partial f_w(S_w)}{\partial x}=0, \tag{3}$$

where $S_w(x,t)$ is water saturation; $v$ denotes the total liquid flow velocity; $f_w(S_w)$ is the water cut, representing the ratio of water flow rate to total flow rate, typically a strongly nonlinear function of saturation with the expression:

$$f_w(S_w)=\frac{1}{1+\dfrac{\mu_w}{\mu_o}\dfrac{k_{ro}(S_w)}{k_{rw}(S_w)}}, \tag{4}$$

where $k_{rw}(S_w)$ and $k_{ro}(S_w)$ are the relative permeabilities of the water and oil phases, respectively, usually described by the Corey model:

$$k_{rw}(S_w)=\begin{cases}0 & S_w\le S_{wc}\\ \left(\dfrac{S_w-S_{wc}}{1-S_{wc}-S_{or}}\right)^{n_w} & S_{wc}<S_w<1-S_{or}\\ 1 & S_w\ge 1-S_{or}\end{cases},\quad k_{ro}(S_w)=\begin{cases}1 & S_w\le S_{wc}\\ \left(\dfrac{1-S_w-S_{or}}{1-S_{wc}-S_{or}}\right)^{n_o} & S_{wc}<S_w<1-S_{or}\\ 0 & S_w\ge 1-S_{or}\end{cases}, \tag{5}$$

in which, $S_{wc}$ is the irreducible water saturation, $S_{or}$ is the residual oil saturation, and $n_w$, $n_o$ are Corey exponents, all determined by core experiments.

The initial condition of the BL equation is generally that the reservoir is saturated with pure oil before waterflooding, and the water saturation equals the irreducible water saturation: $S_w(x,0)=S_{wc}$ for $(x\in[0,L])$, where $L$ is the length of the main streamline in the reservoir. The boundary condition generally adopts a constant saturation boundary at the injection end ($x$= 0): $S_w(0,t)=1-S_{or}$, $(t\in[0,T])$. The core challenge of the BL equation is the shock front induced by nonlinear convection: when the fractional flow function $f_w(S_w)$ is convex, the saturation distribution forms a steep discontinuous front whose position advances with time.

### 2.3 Convection-Diffusion Equation for Concentration in Compositional Flow

Chemical flooding (e.g., polymer flooding) or $CO_2$ sequestration processes in reservoir development involve convection, diffusion, and adsorption of multicomponent substances on rock surfaces. Taking a 2D reservoir domain $\Omega\subset\mathbb{R}^2$ as an example, the convection-diffusion equation for the concentration of a single component (e.g., polymer) is:

$$\phi\frac{\partial c}{\partial t}+\rho_b\frac{\partial q}{\partial t}=\nabla\cdot(D\nabla c)-\nabla\cdot(\mathbf{v}c), \tag{6}$$

where $c=c(\mathbf{x},t)$ denotes the component concentration; $t$ represents time; $\phi$ is the porosity; $\rho_b$ stands for the rock density; $q$ denotes the mass of solute adsorbed per unit mass of rock. It satisfies the linear adsorption law (Henry's law) with the concentration $c$, i.e., $q=K_d c$, where $K_d$ is the distribution coefficient reflecting the rock's adsorption capacity for the component; $D$ represents the diffusion coefficient of the component; $\mathbf{v}=(v_x,v_y)$ denotes the seepage velocity distribution.

The initial condition of the multicomponent convection-diffusion equation is that no injected components are present in the reservoir at the early development stage, with a concentration of zero: $c(\mathbf{x},0)=0,(\mathbf{x}\in\Omega)$. Boundary conditions generally include: a constant concentration injection at the injection boundary $\mathbf{x}\in\Gamma_{\text{in}}$ denoted by $c(\mathbf{x},t)=c_{\text{inj}}$, $(t\in[0,T])$; a convection-dominated free outflow at the production boundary $\mathbf{x}\in\Gamma_{\text{out}}$ denoted by $\nabla c(\mathbf{x},t)\cdot\mathbf{n}=0$, $(t\in[0,T])$; and a closed boundary $\mathbf{x}\in\Gamma_{\text{close}}$ denoted by $\nabla c(\mathbf{x},t)\cdot\mathbf{n}=0$ and $c(\mathbf{x},t)\mathbf{u}(\mathbf{x},t)\cdot\mathbf{n}=0$, $(t\in[0,T])$. In reservoir seepage, the convection velocity is much higher than the diffusion velocity, resulting in convection-dominated equations, sharp concentration fronts, and a high tendency for numerical oscillations.

## 3 Discrete Variable (DV)-Circuit QCPINN Method

The core idea of the DV-Circuit QCPINN (Farea et al., 2025) is to leverage classical networks for input dimension adaptation and feature preprocessing, utilize the superposition/entanglement properties of DV quantum circuits to achieve high-dimensional feature mapping, then employ classical networks again to decode quantum outputs and calculate physical losses, and ultimately optimize both classical and quantum parameters through backpropagation.

Targeting the characteristics of reservoir PDEs, this study adapts this framework to ensure stability and accuracy under scenarios of heterogeneity, strong nonlinearity, and convection dominance. As illustrated in Fig. 1, this framework comprises four components: a classical preprocessor, a DV quantum core, a classical postprocessor, and a physical loss calculation unit. Specifically, the preprocessor maps reservoir inputs (spatiotemporal coordinate parameters of pressure, saturation, or concentration) to quantum input dimensions; the quantum core generates quantum states via parameterized DV circuits and performs measurements; the postprocessor decodes quantum measurement results into physical variables (pressure, saturation, or concentration); and the loss unit computes PDE residual loss as well as boundary/initial condition losses, with all parameters optimized through backpropagation.

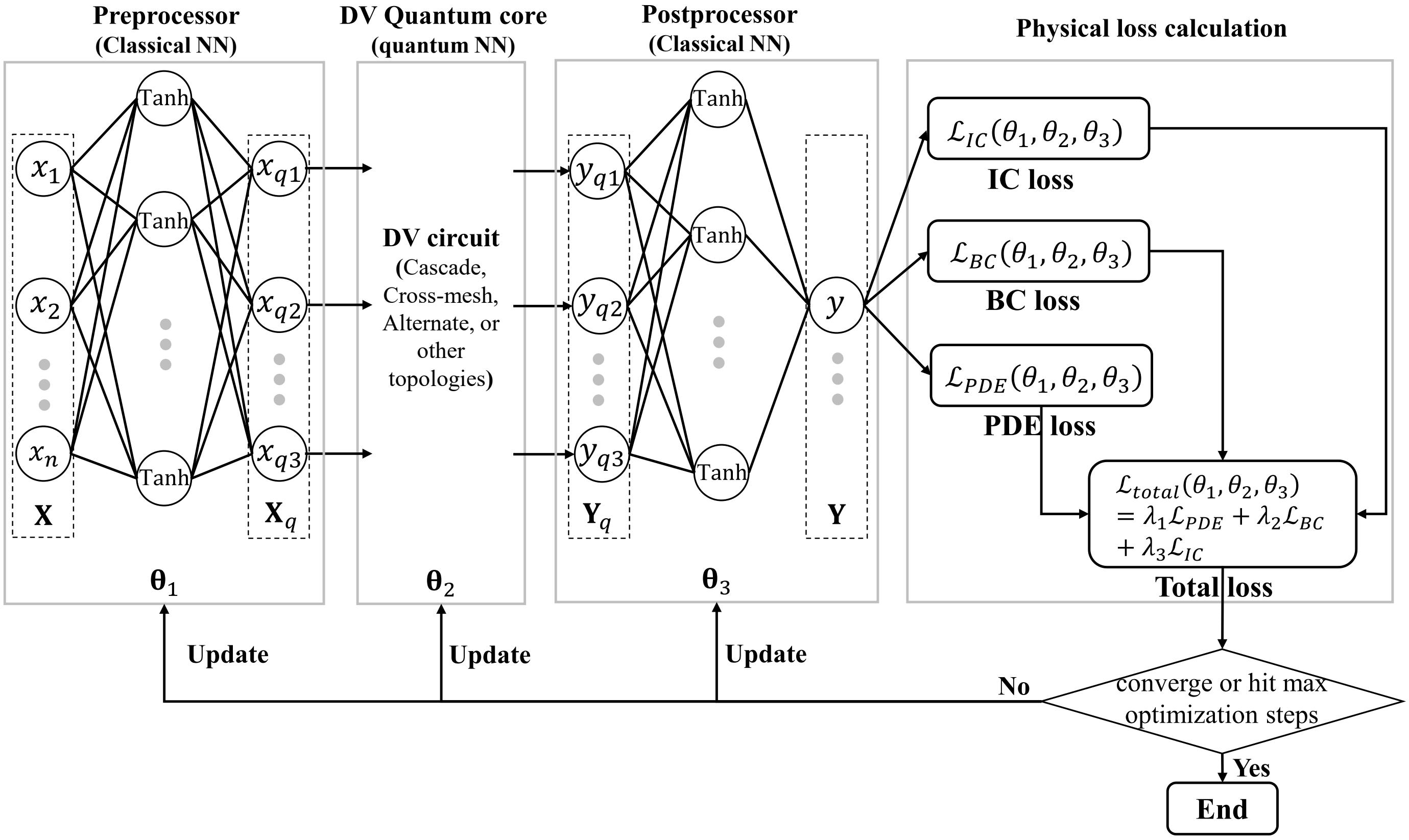

Fig. 1 Schematic diagram of the QCPINN workflow proposed in this study.

Next, we elaborate on the technical details of each component.

### 3.1 Classical preprocessor

The core functions of the classical preprocessor are input dimension adaptation and feature normalization. It converts inputs of reservoir seepage PDEs (e.g., spatiotemporal coordinates $x$, $y$, $t$ for 2D pressure equations) into low-dimensional quantum inputs acceptable to quantum circuits, while reducing sensitivity to parameter initialization. Specifically, it is designed as a fully connected network with a single hidden layer (i.e., one input layer, one hidden layer, and one output layer): the input is the feature vector $\mathbf{X} \in \mathbb{R}^{d_{in}}$ of the reservoir PDE (e.g., for 2D pressure equations, including $x$, $y$), and the number of neurons in the hidden layer is $d_h$. The linear transformation from the input layer to this hidden layer is given by $\mathbf{h}_1 = W_1\mathbf{X} + b_1$, where $W_1 \in \mathbb{R}^{d_h \times d_{in}}$ is the weight matrix, $b_1 \in \mathbb{R}^{d_h}$ is the bias vector, and Xavier initialization is adopted (to avoid gradient vanishing). A Tanh activation function is introduced in the hidden layer to enhance the nonlinear expression capability of the network. Subsequently, the output layer data $\mathbf{X}_q \in \mathbb{R}^{d_q}$ is obtained through full connection between the hidden layer and the output layer (where $d_q$ is the number of qubits used in the quantum circuit), serving as the input to the subsequent quantum circuit, i.e., $\mathbf{X}_q = W_2\mathrm{Tanh}(\mathbf{h}_1) + b_2$. Thus, Eq. (7) presents the computation result of the preprocessor for the feature vector $\mathbf{X}$ of the reservoir PDE. For convenience of description, the set of parameters $W_1$, $W_2$, $b_1$, $b_2$ involved in the classical preprocessor is denoted as $\boldsymbol{\theta}_1$.

$$\mathbf{X}_q = W_2\mathrm{Tanh}(W_1\mathbf{X} + b_1) + b_2 = f_1\left(\mathbf{X}, \boldsymbol{\theta}_1\right), \tag{7}$$

where $f_1$ represents the mapping from the input $\mathbf{X}$ of the preprocessing network and the preprocessing network parameters $\boldsymbol{\theta}_1$ to the input $\mathbf{X}_q$ of the quantum network.

### 3.2 DV Quantum Core

The DV quantum core is the computational engine of QCPINN, responsible for encoding classical preprocessed features into quantum states, achieving high-dimensional feature mapping via Parameterized Quantum Circuits (PQL), and ultimately obtaining classical outputs through measurements.

#### 3.2.1 Quantum State Encoding

This paper adopts the angle-embedding map defined in Eq. (9) to map classical features to quantum states, in which each classical feature independently controls the rotation angle of one qubit, ensuring stable gradient propagation. For quantum input $\mathbf{X}_q = [x_{q1}, x_{q2}, ..., x_{qd_q}]$, each element $x_{qj}$ is treated as a rotation angle. We sequentially apply an RX rotation gate to each qubit $j$ to implement feature encoding, with the unitary transformation of the RX gate defined as:

$$R_X(x_{qj}) = \begin{pmatrix} \cos\left(x_{qj}/2\right) & -i\sin\left(x_{qj}/2\right) \\ -i\sin\left(x_{qj}/2\right) & \cos\left(x_{qj}/2\right) \end{pmatrix}. \tag{8}$$

The initial quantum state of the system is the $d_q$-qubit ground state $|\psi_0\rangle = |0\rangle^{\otimes d_q}$. After applying the RX gates to all qubits, the encoded quantum state is:

$$|\psi_{\text{enc}}\rangle = \bigotimes_{j=1}^{d_q} R_X\left(x_{qj}\right)|0\rangle^{\otimes d_q}. \tag{9}$$

Through local rotation operations, angle embedding directly encodes classical features into the phase of qubits. It should be noted that no CNOT gates are required at all for this state preparation step, because angle encoding relies purely on single qubit rotational gates without any multi-qubit entanglement operations. Entangling gates such as CNOT are only introduced in the subsequent variational quantum circuit layers to generate quantum correlation and enhance feature representation, which aligns with the hardware efficient design suitable for NISQ devices.

Accordingly, the “Angle embedding, Eq. (9)” block in each panel of Fig. 2 denotes this feature-dependent state-preparation unitary.

#### 3.2.2 Parameterized Quantum Circuit (PQC) Topologies

The topology of the quantum circuit determines the entanglement pattern between qubits and directly affects the feature mapping capability. In each subsequent numerical example, three representative parameterized quantum circuit topologies (Cascade, Cross-mesh, and Alternate) are tested. Specifically, the Cascade topology traces its origin to the work of Dallaire-Demers and Killoran (2018); the Cross-mesh topology is derived from the universal quantum circuit framework introduced by Sousa and Ramos (2006); the Alternate topology builds on the device-tailored gate sequence designed by Johnson et al. (2017). These three architectures are systematically collated and validated as typical designs in Sim et al., (2019). Their mathematical formulations and structural details are as follows:

Throughout this subsection, circuit diagrams are read from left to right, whereas matrix products act on state vectors from right to left. An $L$-layer circuit is written as $U^L \cdots U^1$, so layer 1 acts first. The order of every noncommuting gate product is specified explicitly below.

(1) Cascade Topology

Fig. 2(a) shows a single-layer Cascade topology for $d_q$ qubits, adopting a ring connection. Each layer of the circuit consists of two sub-layers: a single-qubit rotation layer and a two-qubit entanglement layer.

Read from left to right in Fig. 2(a), the single-qubit rotation sublayer applies RX with angle $\alpha_j^l$ and then RZ with angle $\beta_j^l$ to qubit $j$. Thus, in rightmost-first matrix notation, its factor is $R_Z\left(\beta_j^l\right)R_X\left(\alpha_j^l\right)$.

$$R_Z(\beta_j^l)=\begin{pmatrix} e^{-i\beta_j^l/2} & 0 \\ 0 & e^{i\beta_j^l/2} \end{pmatrix}. \tag{10}$$

The entanglement layer applies Controlled-RX (CRX) gates along the ring connection. The CRX gate with control qubit $k$ and target qubit $m$ is defined as:

$$CRX\left(\delta_{k,m}^l\right)=|0\rangle\langle 0|\otimes I+|1\rangle\langle 1|\otimes R_X\left(\delta_{k,m}^l\right), \tag{11}$$

where $\delta_{k,m}^l$ is the trainable entanglement parameter, and $I$ is the $2\times 2$ identity matrix.

For $d_q$ qubits and $L$ layers, the total unitary transformation of the Cascade PQC is:

$$U_C\left(\mathbf{\theta}_2\right)=U_C^L\cdots U_C^1,\ U_C^l=\left[CRX\left(\delta_{0,1}^l\right)\cdots CRX\left(\delta_{d_q-1,0}^l\right)\right]\left[\bigotimes_{j=0}^{d_q-1}R_Z\left(\beta_j^l\right)R_X\left(\alpha_j^l\right)\right], \tag{12}$$

in which, the directed ring is $0\to 1,\ 1\to 2,\ ...,\ (n-1)\to 0$. The CRX factors are written in that algebraic order, with the rightmost factor acting first. For $d_q$ = 3, Fig. 2(a) therefore applies $CRX\left(\delta_{2,0}^1\right)$, $CRX\left(\delta_{1,2}^1\right)$, and $CRX\left(\delta_{0,1}^1\right)$ chronologically. Because CRX gates sharing a qubit do not generally commute, this order is part of the circuit definition. The vector $\mathbf{\theta}_2$ collects all PQC trainable parameters.

The PQC in Fig. 2(a) shows $d_q=3$ qubits and $L$=1 layer (a constraint of NISQ-era hardware), the circuit depth is $\left(d_q+2\right)L=5$, and the number of parameters is $3\times d_q\times L=9$, including $2d_q\times L=6$ single-qubit parameters and $d_q\times L=3$ entanglement parameters, far fewer than the thousands of parameters in classical PINNs.

(2) Cross-mesh topology

Fig. 2(b) depicts a single-layer Cross-mesh topology for $d_q$ qubits, featuring a fully connected entanglement structure (each qubit is entangled with all other qubits). The circuit consists of three sub-layers in sequence: a pre-single-qubit rotation layer, a fully connected entanglement layer, and a post-single-qubit rotation layer.

Read from left to right in Fig. 2(b), the pre-rotation sublayer applies RX with angle $\zeta_j^l$ and then RZ with angle $\theta_j^l$ to qubit $j$; its rightmost-first matrix factor is $R_Z\left(\theta_j^l\right)R_X\left(\zeta_j^l\right)$.

The entanglement sublayer applies Controlled-RZ (CRZ) gates to every directed qubit pair ($k$, $m$) with $k$ different from $m$. Each CRZ gate is diagonal under the computational basis. Therefore, all CRZ factors commute, including factors that share a qubit. Their order does not change the unitary. For reproducibility, directed pairs are enumerated lexicographically in the implementation and in Fig. 2(b). The CRZ gate with control qubit $k$ and target qubit $m$ is defined as:

$$\mathrm{CRZ}\left(\delta_{k,m}^l\right)=|0\rangle\langle 0|\otimes I+|1\rangle\langle 1|\otimes R_Z\left(\delta_{k,m}^l\right). \tag{13}$$

The post-rotation sublayer applies RX with angle $\gamma_j^l$ and then RZ with angle $\eta_j^l$ to qubit $j$. Its rightmost-first matrix factor is $R_Z\left(\eta_j^l\right)R_X\left(\gamma_j^l\right)$.

The total unitary transformation of the Cross-mesh PQC is:

$$U_X\left(\mathbf{\theta}_2\right)=U_X^L\cdots U_X^1,$$
$$U_X^l=\left[\bigotimes_{j=0}^{d_q-1}R_Z\left(\eta_j^l\right)R_X\left(\gamma_j^l\right)\right]\left[\prod_{k\neq m}CRZ\left(\delta_{k,m}^l\right)\right]\left[\bigotimes_{j=0}^{d_q-1}R_Z\left(\theta_j^l\right)R_X\left(\zeta_j^l\right)\right]. \tag{14}$$

The PQC in Fig. 2(b) shows $d_q$=3 and $L$=1, the circuit depth is $\left(d_q^{\,2}-d_q+4\right)L=10$ layers, and the number of parameters is $\left(d_q^{\,2}+3d_q\right)L=18$, including $4d_q\times L=12$ single-qubit parameters and $d_q\left(d_q-1\right)L=6$ entanglement parameters.

(3) Alternate topology

Fig. 2(c) illustrates the Alternate topology for $d_q$ qubits. Let $E_0=\{(0,1),(2,3),\ldots\}$ be the even-bond set. For odd $d_q$, let $E_1=\{(1,2),(3,4),\ldots\}$; for even $d_q$, $E_1$ additionally contains the periodic closing bond $\left(d_q-1,0\right)$. Let $V_0$ and $V_1$ denote the qubits incident to $E_0$ and $E_1$, respectively. A layer $l$ consists of the following six chronological sublayers:

Step 1: $A_l$ applies RY with trainable angle $\alpha_j^l$ to every qubit $j=0,1,\ldots,d_q-1$, which is expressed in Eq. (15).

$$A_l=\bigotimes_{j=0}^{d_q-1}R_Y\left(\alpha_j^l\right), \tag{15}$$

in which, the RY matrix is given in Eq. (16).

$$R_Y\left(\alpha_j^l\right)=\begin{pmatrix}\cos\left(\alpha_j^l/2\right) & -\sin\left(\alpha_j^l/2\right)\\ \sin\left(\alpha_j^l/2\right) & \cos\left(\alpha_j^l/2\right)\end{pmatrix}. \tag{16}$$

Step 2: $C_{0,l}$ applies CNOT gate in parallel for every bond $\left(j,j+1\right)$ in $E_0$. These bonds are disjoint, so their gates commute within the sublayer.

$$C_{0,l}=\prod_{(j,j+1)\in E_0}CNOT\left(\delta_{j,j+1}^l\right), \tag{17}$$

where $\mathrm{CNOT}=|0\rangle\langle 0|\otimes I+|1\rangle\langle 1|\otimes X$, $X$ is the Pauli-X gate, $X=\begin{pmatrix}0 & 1\\ 1 & 0\end{pmatrix}$.

Step 3: $B_l$ applies RZ with trainable angle $\beta_j^l$ to every $j$ in $V_0$. RZ matrix is defined in Eq. (10).

$$B_l=\prod_{j\in V_0}R_Z\left(\beta_j^l\right). \tag{18}$$

Step 4: $G_l$ applies RY with trainable angle $\gamma_j^l$ to every qubit common to $V_0$ and $V_1$.

$$G_l=\prod_{j\in V_0\cap V_1}R_Y\left(\gamma_j^l\right). \tag{19}$$

Step 5: $C_{1,l}$ applies CNOT gate in parallel for every bond $\left(j,j+1\right)$ in $E_1$:

$$C_{1,l}=\prod_{(j,j+1)\in E_1}CNOT\left(\delta_{j,j+1}^l\right). \tag{20}$$

Step 6: $D_l$ applies RZ with trainable angle $\zeta_j^l$ to every $j$ in $V_1$.

$$D_l=\prod_{j\in V_1}R_Z\left(\zeta_j^l\right) \tag{21}$$

With the rightmost sublayer acting first, the Alternate circuit is:

$$U_A(\theta_2)=U_A^L\cdots U_A^1,\ \ U_A^l=D_lC_{1,l}G_lB_lC_{0,l}A_l. \tag{22}$$

For $d_q$=3 and $L$=1, this definition gives exactly Fig. 2(c): RY on qubits 0-2, $CNOT\left(\delta_{0,1}^1\right)$, RZ on qubits 0-1, RY on qubit 1, $CNOT\left(\delta_{1,2}^1\right)$, and RZ on qubits 1-2. For odd $d_q$, each layer contains $d_q$-1 CNOT gates and $4\left(d_q-1\right)$ trainable rotation angles. For even $d_q$, the periodic closing bond gives $d_q$ CNOT gates and $4d_q$ trainable rotation angles per layer. Thus, the corresponding $L$-layer counts are $\left(d_q-1\right)L$ and $4\left(d_q-1\right)L$ for odd $d_q$, or $d_qL$ and $4d_qL$ for even $d_q$; the depth is $6L$ in both cases.

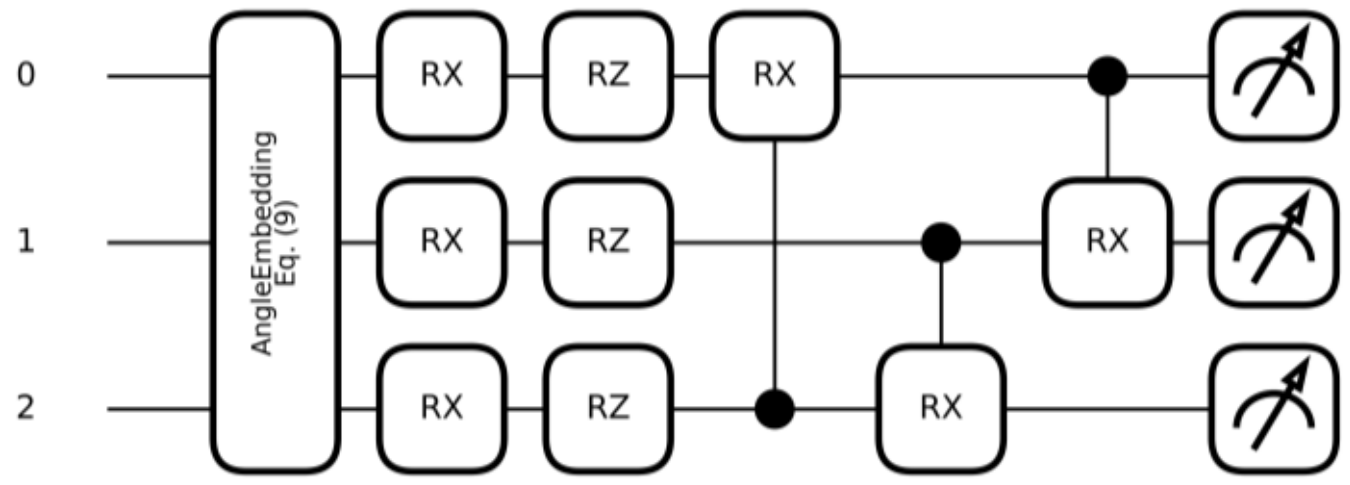

(a) Cascade

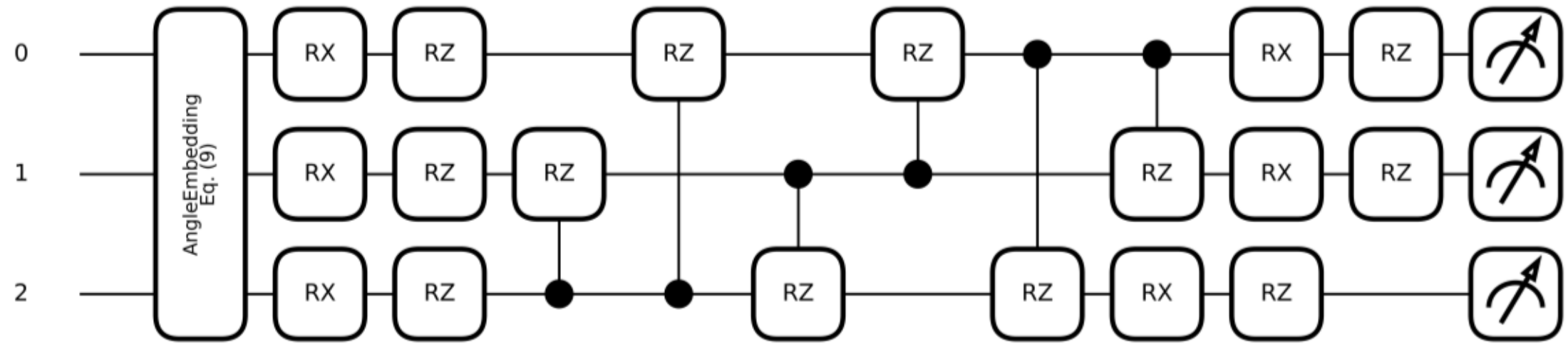

(b) Cross-mesh

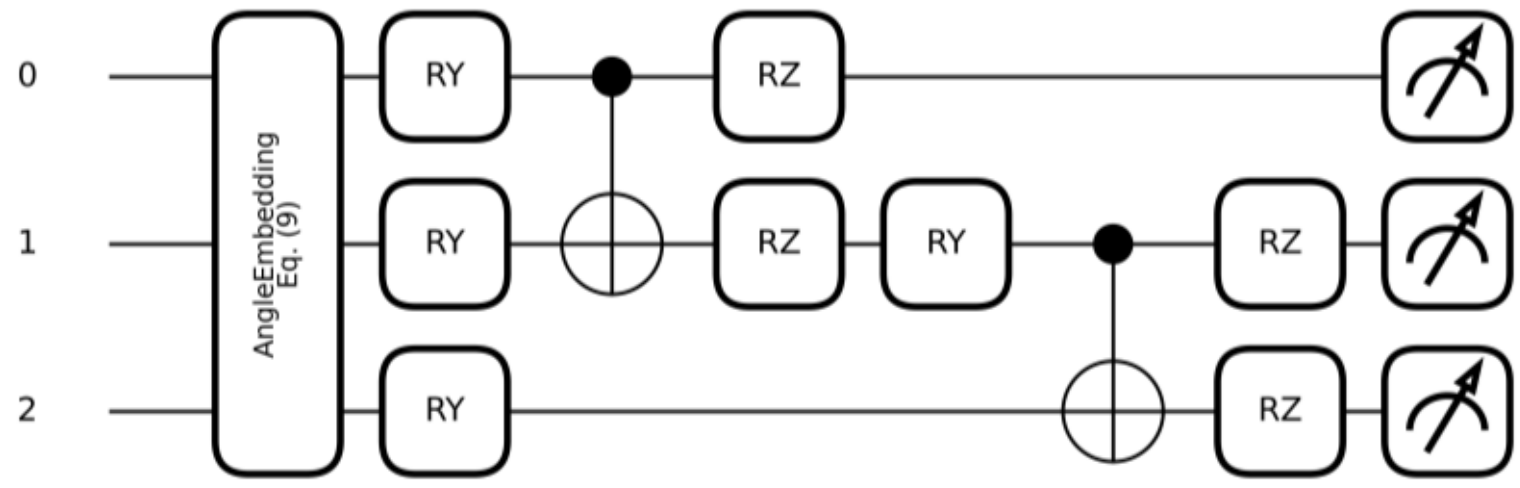

(c) Alternate

Fig. 2 Three quantum-circuit topologies used in this study. In all panels, the Angle embedding block implements Eq. (9), and gates are applied from left to right.

Each topology follows a systematic and mathematically defined construction rule. Given the number of qubits $d_q$ and the number of quantum layers $L$, the gate positions, types, parameter counts, and entanglement patterns are uniquely determined. The degrees of freedom (DOF) increase quantitatively and predictably with $d_q$ and $L$, ensuring well-defined scalability consistent with the classical network. This unified scaling mechanism ensures that the QCPINN framework can be systematically expanded to higher representation capacity.

### 3.2.3 Quantum Measurement

Quantum measurement is a key step in mapping quantum states to classical outputs. This paper adopts Pauli-Z expectation measurement: for each qubit $i$, the expected value $\langle Z_i \rangle$ under the Pauli-Z basis is measured, where $Z$ is the Pauli-Z operator ($Z=\begin{pmatrix}1 & 0\\ 0 & -1\end{pmatrix}$). The expectation value is calculated as:

$$\langle Z_i \rangle = \langle \psi_{\text{final}} | Z_i \otimes I^{\otimes(n-1)} | \psi_{\text{final}} \rangle, \tag{23}$$

where $|\psi_{\text{final}}\rangle = U(\boldsymbol{\theta}_2)|\psi_{\text{enc}}\rangle$ is the quantum state after passing through the PQC, and $U\left(\boldsymbol{\theta}_2\right)$ is the total unitary transformation of the PQC, i.e., Eqs. (12), (14), or (21) depending on the topology.
The measurement results form a classical vector $\mathbf{Y}_q = [\langle Z_1\rangle, \langle Z_2\rangle, ..., \langle Z_n\rangle]^T$, where each element ranges from $\left[-1,\ 1\right]$. This vector is then input to the classical postprocessor for physical variable mapping.

The quantum core can be abstractly expressed as:

$$\mathbf{Y}_q = f_2\left(\mathbf{X}_q, \boldsymbol{\theta}_2\right), \tag{24}$$

where $f_2(\cdot)$ represents the mapping from the quantum input $\mathbf{X}_q$ and quantum parameters $\boldsymbol{\theta}_2$ to the quantum output $\mathbf{Y}_q \in \mathbb{R}^{d_q}$.

### 3.3 Classical Postprocessor

The classical postprocessor is responsible for quantum output decoding and physical variable mapping, converting the low-dimensional vector from quantum measurements into the target physical variables of reservoir seepage PDEs (pressure *p*, saturation $S_w$, or component concentration *c*). It is designed as a fully connected network with a single hidden layer, symmetric to the preprocessor: the input of the first layer (quantum output decoding) is the quantum measurement vector $\mathbf{Y}_q \in \mathbb{R}^{d_q}$, which is linearly transformed to the hidden layer vector $\mathbf{h}_3 \in \mathbb{R}^{d_h}$, i.e., $\mathbf{h}_3 = W_3\mathbf{Y}_q + b_3$, where $W_3 \in \mathbb{R}^{d_h \times d_q}$ and $b_3 \in \mathbb{R}^{d_h}$ are trainable parameters initialized via Xavier. The Tanh activation function is introduced in this hidden layer, followed by a linear transformation through full connection to obtain the output layer data $\mathbf{Y}$. Thus, Eq. (9) summarizes the operation process of the classical postprocessor. For convenience, the set of parameters $W_3$, $W_4$, $b_3$, $b_4$ involved in the classical postprocessor is denoted as $\boldsymbol{\theta}_3$.

$$\mathbf{Y} = W_4 \mathrm{Tanh}\left(W_3\mathbf{Y}_q + b_3\right) + b_4 = f_3\left(\mathbf{Y}_q, \boldsymbol{\theta}_3\right), \tag{25}$$

where $W_4 \in \mathbb{R}^{1 \times d_h}$ and $b_3 \in \mathbb{R}^1$. $f_3$ represents the mapping from the output $\mathbf{Y}_q$ of the quantum network and the postprocessing network parameters $\boldsymbol{\theta}_3$ to the output $\mathbf{Y}$ of the postprocessing network.

Combining Eqs. (7), (8), and (9), the mapping of input $\mathbf{X} \in \mathbb{R}^{d_{in}}$ to output $\mathbf{Y} \in \mathbb{R}^1$ for the hybrid quantum-classical network in Fig. 1 can be denoted as:

$$\mathbf{Y} = f\left(\mathbf{X}, \boldsymbol{\theta}_1, \boldsymbol{\theta}_2, \boldsymbol{\theta}_3\right). \tag{26}$$

Physical loss is the core guarantee of QCPINN's physical consistency requiring simultaneous minimization of PDE residual loss (to ensure the solution satisfies the governing equation) and boundary/initial condition losses (to ensure the solution satisfies constraints). The total loss function is:

$$\mathcal{L}_{total}\left(\boldsymbol{\theta}_1, \boldsymbol{\theta}_2, \boldsymbol{\theta}_3\right) = \lambda_1 \mathcal{L}_{PDE}\left(\boldsymbol{\theta}_1, \boldsymbol{\theta}_2, \boldsymbol{\theta}_3\right) + \lambda_2 \mathcal{L}_{BC}\left(\boldsymbol{\theta}_1, \boldsymbol{\theta}_2, \boldsymbol{\theta}_3\right) + \lambda_3 \mathcal{L}_{IC}\left(\boldsymbol{\theta}_1, \boldsymbol{\theta}_2, \boldsymbol{\theta}_3\right), \tag{27}$$

where $\lambda_1$, $\lambda_2$, and $\lambda_3$ are loss weights (we set $\lambda_1 = 1$, $\lambda_2 = 1$, $\lambda_3 = 1$ in this paper). The PDE residual loss is calculated using Mean Squared Error (MSE) for the residuals of the PDE governing equation based on automatic differentiation at sampling points. The boundary condition loss computes the MSE between predicted values and boundary conditions at boundary sampling points, covering both Dirichlet and Neumann boundaries. The initial condition loss calculates the MSE between predicted values and initial conditions at sampling points.

The parameters of the DV-circuit QCPINN include classical parameters $\boldsymbol{\theta}_1$ and $\boldsymbol{\theta}_3$ as well as quantum parameters $\boldsymbol{\theta}_2$. The optimization process updates both types of parameters simultaneously. The Adam optimizer is adopted, combined with a ReduceLROnPlateau scheduler to dynamically decay the learning rate during loss plateaus. This reduces sensitivity to quantum parameter initialization and helps escape local optima. Additionally, gradient clipping is applied to limit the gradient norm, preventing gradient explosion of parameter gradients.

The DV-circuit QCPINN code in this paper is developed based on the open-source code by Farea et al. (2025) and the PennyLane platform. Key modifications and adaptations include reservoir adjustments specific to the PDE system and tailored designs of the DV quantum circuit topology. In our implementation, PennyLane is configured with shots=None to enable analytical gradient computation, thereby avoiding gradient fluctuations induced by sampling noise. For an ideal quantum processor without hardware errors, the number of shots required to sufficiently suppress gradient noise primarily depends on the qubit count and acceptable error tolerance, following a general scaling law widely observed in variational quantum algorithms (Schuld et al., 2019). For the small qubit systems employed in this work, several thousand to a few tens of thousands of shots are sufficient to reduce gradient noise to a level comparable to our analytical simulations (Seksaria and Prabhakar, 2025). Such scaling behavior is manageable within the capacity of near-term quantum hardware, supporting the practical deployment of the proposed QCPINN method in realistic quantum computing environments.

Furthermore, from a theoretical perspective, the parameterized quantum circuits in this study satisfy the universal approximation theorem (Goto and Miyamoto, 2021; Gonon and Jacquier, 2023). With sufficient qubits and layers, the model can approximate any continuous function to arbitrary precision, and the error decreases toward zero as DOF increases. However, in practical numerical experiments, we do not observe monotonic error reduction with increasing $d_q$ or $L$. This discrepancy arises from two fundamental challenges in variational quantum algorithms. First, barren plateaus induce exponentially vanishing gradients, not only with increasing circuit depth (McClean et al., 2018) but also with increasing qubit count even in shallow circuits (Cerezo et al., 2021b). Second, the optimization landscape contains numerous low-quality local minima, which trap the solver even when barren plateaus are absent (Anschuetz and Kiani, 2022). These effects prevent ideal convergence behavior despite the theoretical universal approximation guarantee.

## 4 Numerical Experiments

### 4.1 Example 1

This example focuses on the heterogeneous-permeability single-phase flow problem described in Section 2.1. Consider a 2D rectangular reservoir $\Omega = [0,100]m \times [0,100]m$ . By setting $k(\mathbf{x}) = K_0 + y$, $K_0 = 100mD$, $\mu = 1cp$, and $q_s(\mathbf{x}) = 0$ in Eq. (2). As shown in Fig. 3, the upper and lower boundaries of the reservoir in this example are constant-pressure boundaries, with the upper boundary pressure being 5 MPa and the lower boundary pressure 10 MPa. The left and right boundaries are set as impermeable closed boundaries.

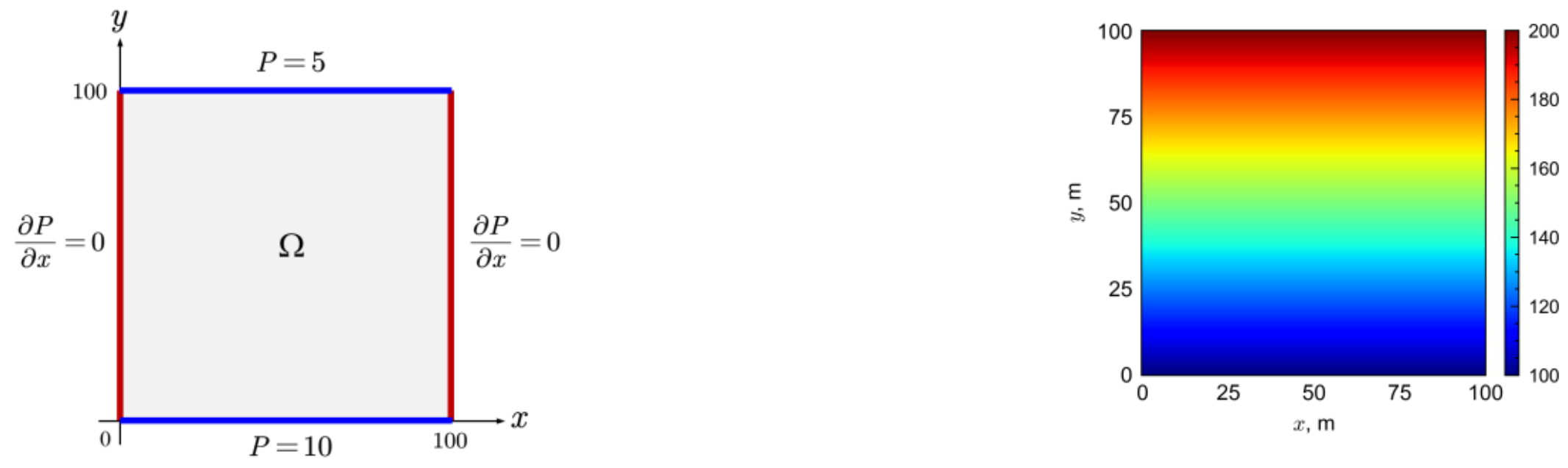

(a) Reservoir domain and boundary conditions (b) Heterogeneous permeability distribution

Fig. 3 Schematic of boundary conditions for the reservoir domain and heterogeneous permeability distribution in Example 1.

The QCPINN model in this example takes spatial coordinates $(x, y)$ as input and outputs the pressure value $p$. Table 1 lists the detailed parameters and training configurations of the QCPINN model (classical network and DV quantum circuit) used in this example. Here, the batch size denotes the number of collocation points sampled from each loss sampler per epoch. In Example 1, 64 points are sampled for each of the four boundary-condition terms and the interior PDE residual term, giving 320 collocation points per epoch. These points are randomly resampled at each epoch rather than fixed throughout training, which helps reduce the bias from a static collocation grid and improves the enforcement of the physical constraints over the whole solution domain.

To systematically evaluate the impact of different quantum circuit topologies on QCPINN performance,

this study adopted three quantum circuit topologies: Cascade, Cross-mesh, and Alternate, as shown in Fig. 4. During training, the Adam optimizer was used with an initial learning rate of 0.005. The ReduceLROnPlateau scheduler was introduced to reduce the learning rate when the loss showed no improvement for multiple consecutive epochs. The patience value was set to 1000 epochs, the learning rate decay factor was 0.9 (a 10% reduction each time), and the minimum learning rate was $10^{-6}$ (to prevent training stagnation due to excessively low learning rates). Meanwhile, the maximum parameter gradient norm was set to 1.0, and gradient clipping was applied to limit the gradient norm. To highlight the performance advantages of QCPINN, this example introduces a comparative analysis between the PINN solution and the QCPINN solution. To ensure fairness in the comparison, the PINN adopts a parameter configuration equivalent to that of QCPINN: the neural network architecture of the PINN is consistent with that of QCPINN in terms of the input and output layers, as well as the first and last hidden layers, each containing 20 neurons. The middle-hidden layer is set to 2 neurons, resulting in 183 trainable parameters for the PINN model, which is comparable to the 189 (Cascade), 193 (Cross-mesh), and 191 (Alternate) trainable parameters for three QCPINN models (including those of the preprocessing layer, quantum layer, and postprocessing layer). Meanwhile, the activation function, initial learning rate, batch size, optimizer type, and maximum number of training epochs of the PINN are exactly the same as those of QCPINN, eliminating the interference of training configuration differences on performance evaluation. It should be noted that all numerical experiments were carried out on a GPU platform equipped with an AMD Ryzen 9 9955HX3D 16-core CPU and an NVIDIA RTX 5070i GPU. The GPU substantially accelerates the quantum state vector evolution and analytical gradient calculations, which dominate the computational cost of QCPINNs. Given the small circuit scale (2 qubits) and batch size of 64, this configuration ensures stable and efficient training.

Fig. 5 shows the training loss evolution of the three QCPINN topologies and the classical PINN. All models decay rapidly from an initial loss on the order of $10^1$. The Cross-mesh QCPINN converges the fastest and achieves the lowest final loss on the order of $10^{-5}$, followed by the Alternate topology at $10^{-4}$. The Cascade QCPINN and PINN stabilize around $10^{-3}$, with the PINN showing more persistent fluctuations than the QCPINNs. Overall, all QCPINN variants outperform the classical PINN in both convergence speed and final loss, with Cross-mesh delivering the most robust performance.

Since the pressure gradient in the x-direction can be shown to be zero in this example, the governing PDE reduces to an ordinary differential equation (ODE) in the y-direction. The analytical solution can thus be readily derived with the given boundary conditions and used as the reference solution. Fig. 6 compares the reference solution, the PINN prediction results, and the QCPINN prediction results with the three quantum circuit topologies, and further shows the corresponding absolute error distributions. Visually, all models achieve pressure distribution solutions that are highly consistent with the reference solution. Compared with the QCPINNs with the three topologies, the deviation between the PINN and the reference solution is more significant, especially near the corner regions, indicating that the accuracy of the PINN is significantly lower than that of the QCPINNs. The quantitative error metrics are detailed in Table 2 and visualized in Fig. 7. The PINN exhibits the highest errors across all metrics, confirming its inferior performance compared to all three QCPINN variants. Among the QCPINN models, the Cross-mesh topology achieves the best accuracy, with the lowest Mean Absolute Error (MAE) 0.001313 and Mean Relative Error (MRE) 0.0198%, followed by the Alternate topology (MAE=0.001466) and the Cascade topology (MAE=0.002742). This ranking is consistent across all error metrics, demonstrating that the full-connectivity entanglement of the Cross-mesh topology is most effective at capturing the global coupling characteristics of the elliptic pressure diffusion problem.

In summary, for the heterogeneous reservoir seepage problem governed by the elliptic equation in this example, all three QCPINN variants significantly outperform the traditional PINN in terms of accuracy. Among them, the Cross-mesh topology achieves the highest computational accuracy, followed by the Alternate and Cascade topologies. These results fully verify that QCPINNs integrated with quantum circuits offer superior stability and accuracy compared with traditional PINNs for solving single-phase flow problems in heterogeneous reservoirs.

As shown in Table 3, on the GPU-accelerated platform, the classical PINN achieves the shortest training time of 89.252s, while the QCPINNs with different topologies take 200.507~246.815s. The longer runtime of QCPINNs arises from the inherent overhead of emulating quantum superposition and entanglement on classical hardware, even though GPU acceleration has already greatly narrowed the runtime gap compared with CPU-based simulations. This confirms that QCPINNs deliver superior accuracy at the cost of moderate additional training time on classical GPUs, and their runtime advantage will be fully unlocked on actual near-term quantum hardware.

To further investigate the scalability behavior of the Cross-mesh QCPINN with respect to the number of qubits and circuit layers, we performed a systematic sensitivity analysis, as illustrated in Fig. 8. For a fixed circuit depth of 1 layer, the MAE decreases monotonically as the number of qubits increases from 2 to 8, indicating that model expressiveness and accuracy improve consistently with increasing quantum degrees of freedom. This trend aligns well with the theoretical universal approximation property.

In contrast, when the number of qubits is fixed at 2, the MAE remains relatively stable from 1 to 3 layers but increases noticeably at 4 layers, indicating that increasing circuit depth does not necessarily lead to monotonic accuracy improvement. To further examine whether this behavior is mainly caused by the sampling budget, we repeated the layer-sensitivity test using enlarged batch sizes of 1024, corresponding to 5120 randomly resampled collocation points per epoch. As shown in Fig. 8(c), increasing the batch size generally reduces the MAE compared with the batch size of 64, suggesting that more collocation points strengthen the physical supervision and reduce sampling-related error. However, the 4-layer model still produces a larger MAE than the 3-layer model in this enlarged-batch experiment. This persistent non-monotonic behavior suggests that deeper parameterized circuits may introduce a more challenging optimization landscape, increased parameter redundancy, and possible barren-plateau-like gradient attenuation. Therefore, improving QCPINN performance requires not only increasing circuit expressiveness or sampling budget, but also maintaining trainability as the circuit depth increases.

To further evaluate the influence of finite quantum measurements, we conducted a shots-sensitivity experiment using the trained Cross-mesh QCPINN in Fig. 9. The trainable parameters were kept fixed as Table 1, and only the number of measurement shots was changed during inference. The analytical simulator results with shots=None was used as the reference baseline. As the number of shots increases from 100 to 100000, both the MAE and the relative L2 error decrease monotonically. These results indicate that the additional inference error introduced by finite shots is mainly caused by measurement sampling uncertainty. Therefore, the proposed QCPINN framework can be used to benchmark quantum hardware by comparing the measured finite-shot accuracy against the ideal analytical-inference baseline and the expected shot-noise scaling.

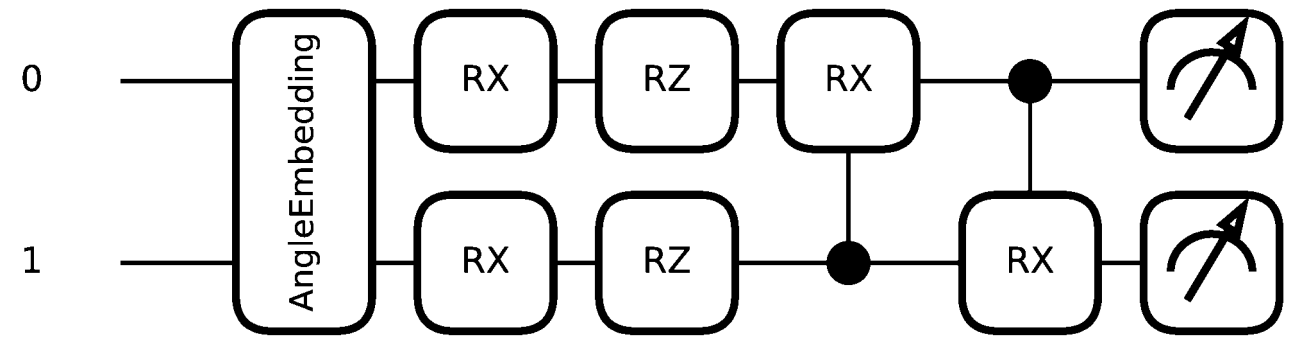

(a) Cascade

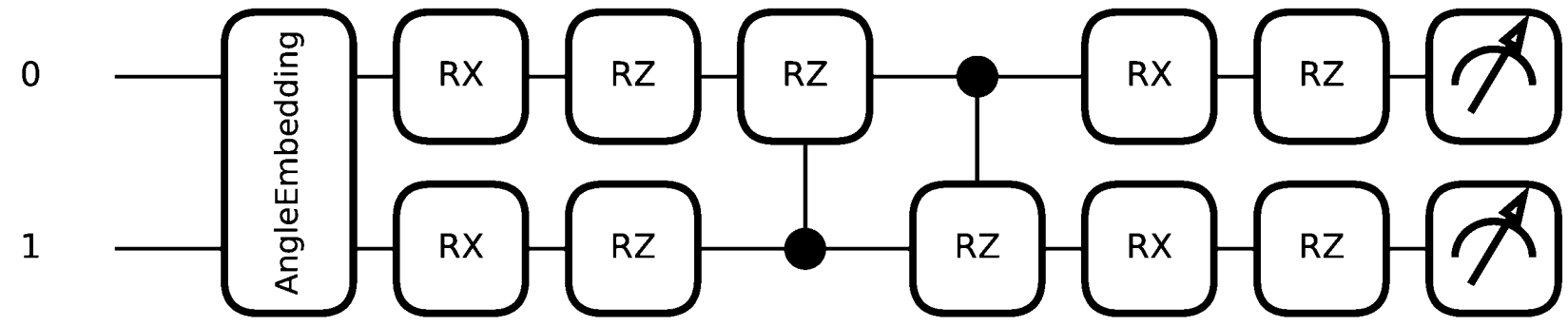

(b) Cross-mesh

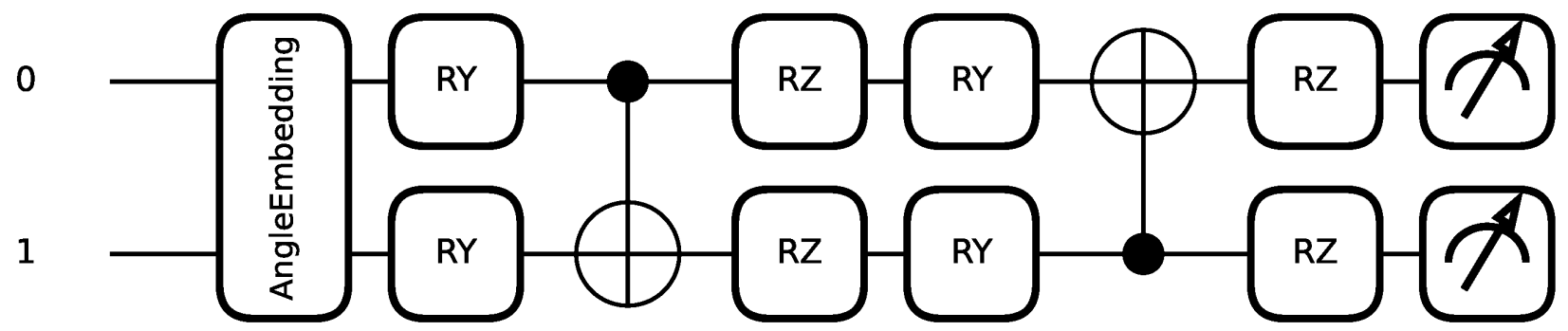

(c) Alternate

Fig. 4 Three quantum circuit topologies used in Example 1.

Table 1 Parameters of the neural networks and optimizers in Example 1.

| Configuration Category | Parameter Name | Setting Value |
|---|---|---|
| Neural Network | Hidden Layers | 1 layer with 20 neurons |
| | Activation Function | Tanh |
| DV Quantum circuit | Circuit Topology | Cascade |
| | | Cross-mesh |
| | | Alternate |
| | Number of Qubits | 2 |
| | Number of Quantum Layers | 1 |
| | Encoding | angle |
| Training Hyperparameters | Optimizer | Adam |
| | Initial Learning Rate | 0.005 |
| | Batch Size | 64 |
| | Maximum Training Epochs | 20000 |

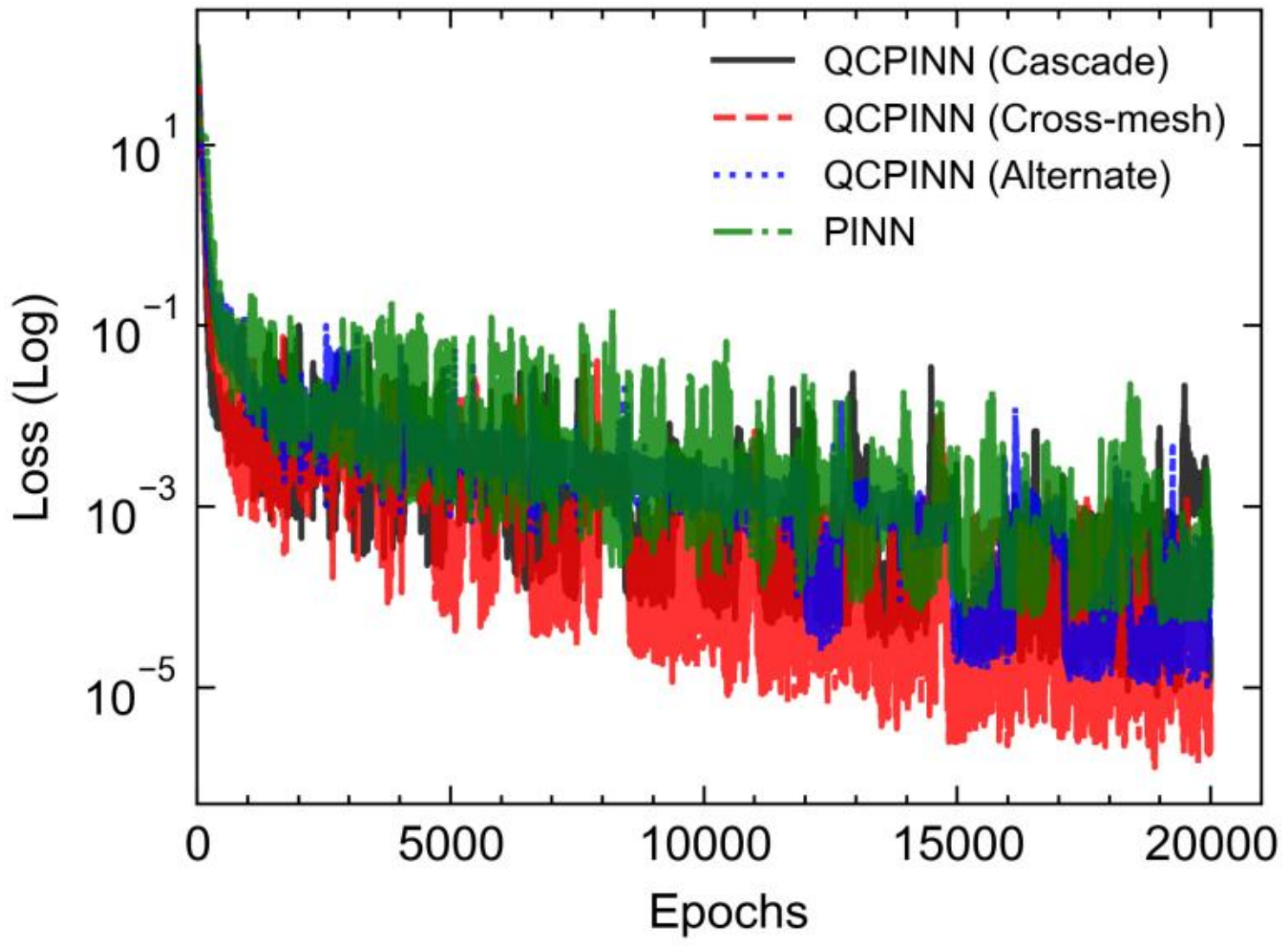

Fig. 5 Comparison of loss function versus epochs of QCPINN with three quantum circuit topologies and PINN in Example 1.

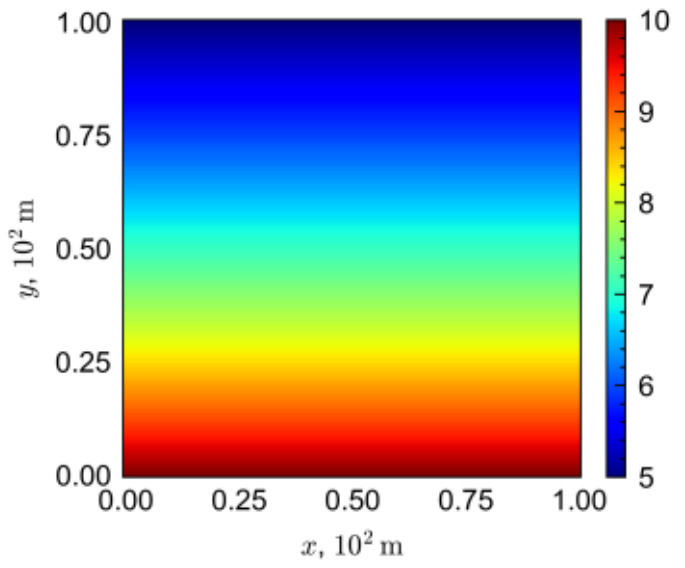

Reference solution

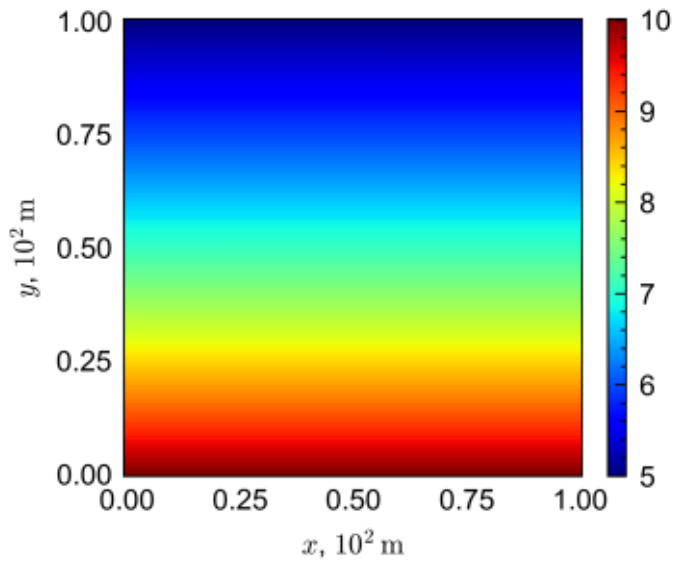

PINN solution

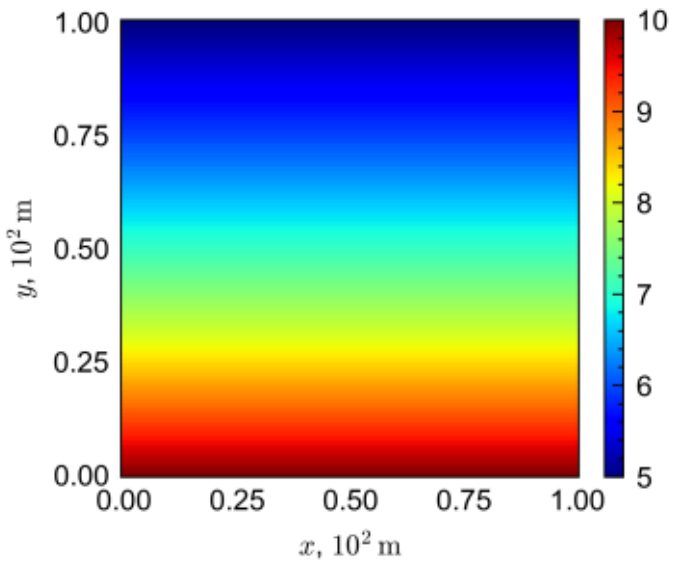

QCPINN solution (Cascade)

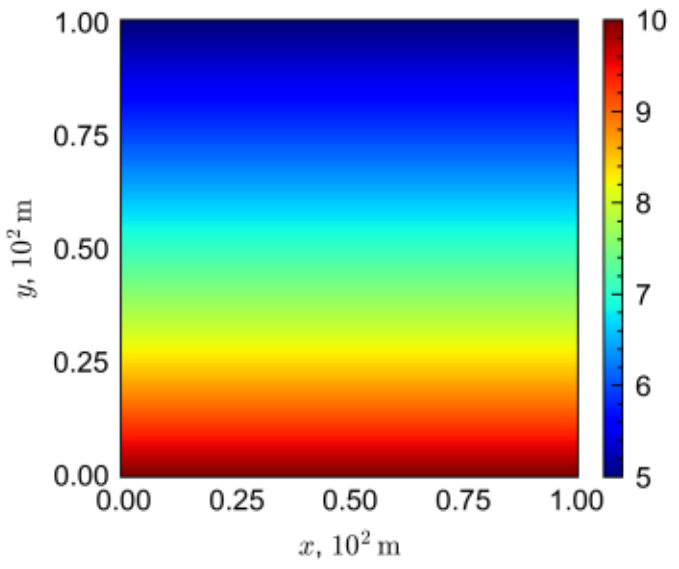

QCPINN solution (Cross-mesh)

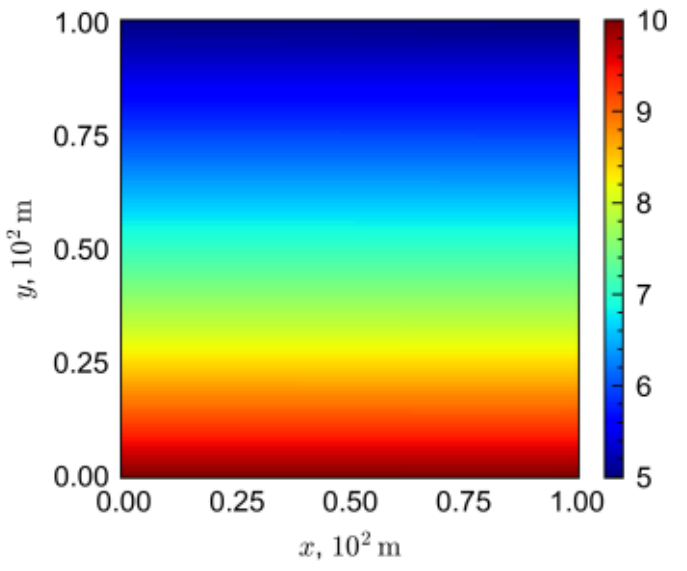

QCPINN solution (Alternate)

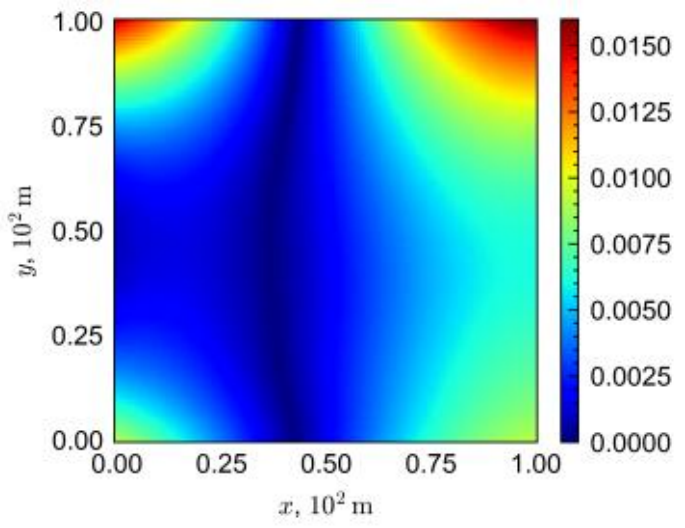

Error distribution for PINN

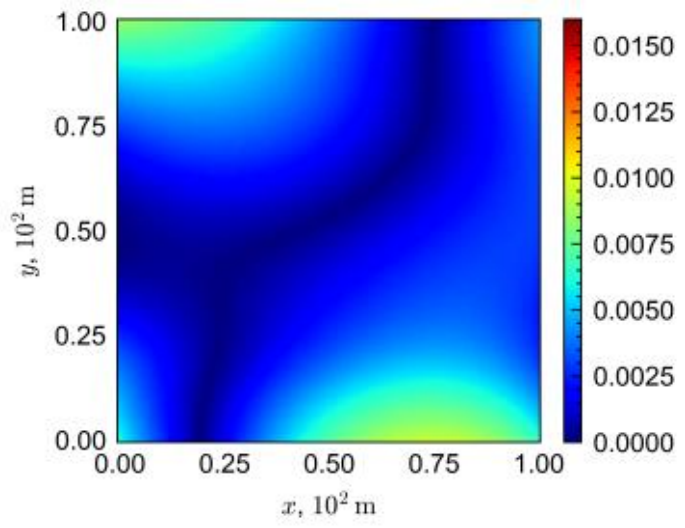

Error distribution for QCPINN (Cascade)

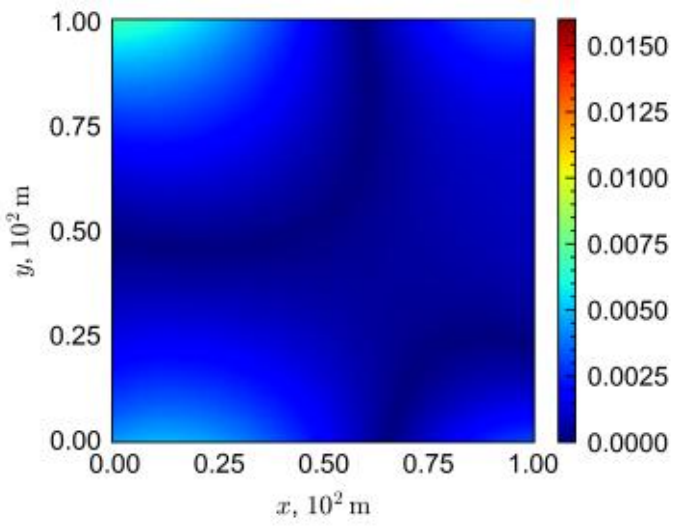

Error distribution of QCPINN (Cross-mesh)

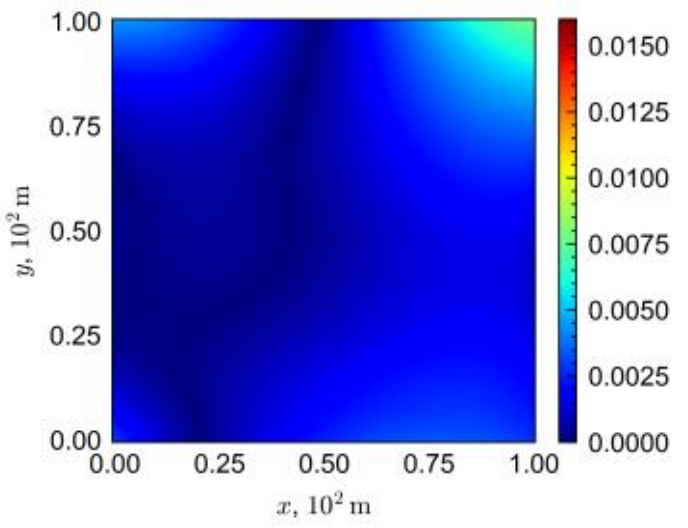

Error distribution of QCPINN (Alternate)

Fig. 6 Comparison of pressure distributions and absolute error distributions of QCPINN predictions using three topologies and PINN prediction in Example 1.

Table 2 Comparison of errors of the pressure profiles: PINN vs QCPINN using three quantum circuit topologies in Example 1.

| Method | Mean Absolute Error | Max Absolute Error | Mean Relative Error | Max Relative Error |
|---|---|---|---|---|
| QCPINN (Cascade) | 0.002742 | 0.009417 | 0.0391% | 0.1698% |
| QCPINN (Cross-mesh) | 0.001313 | 0.006753 | 0.0198% | 0.1351% |
| QCPINN (Alternate) | 0.001466 | 0.007903 | 0.02237% | 0.1581% |
| PINN | 0.003391 | 0.016964 | 0.0530% | 0.3393% |

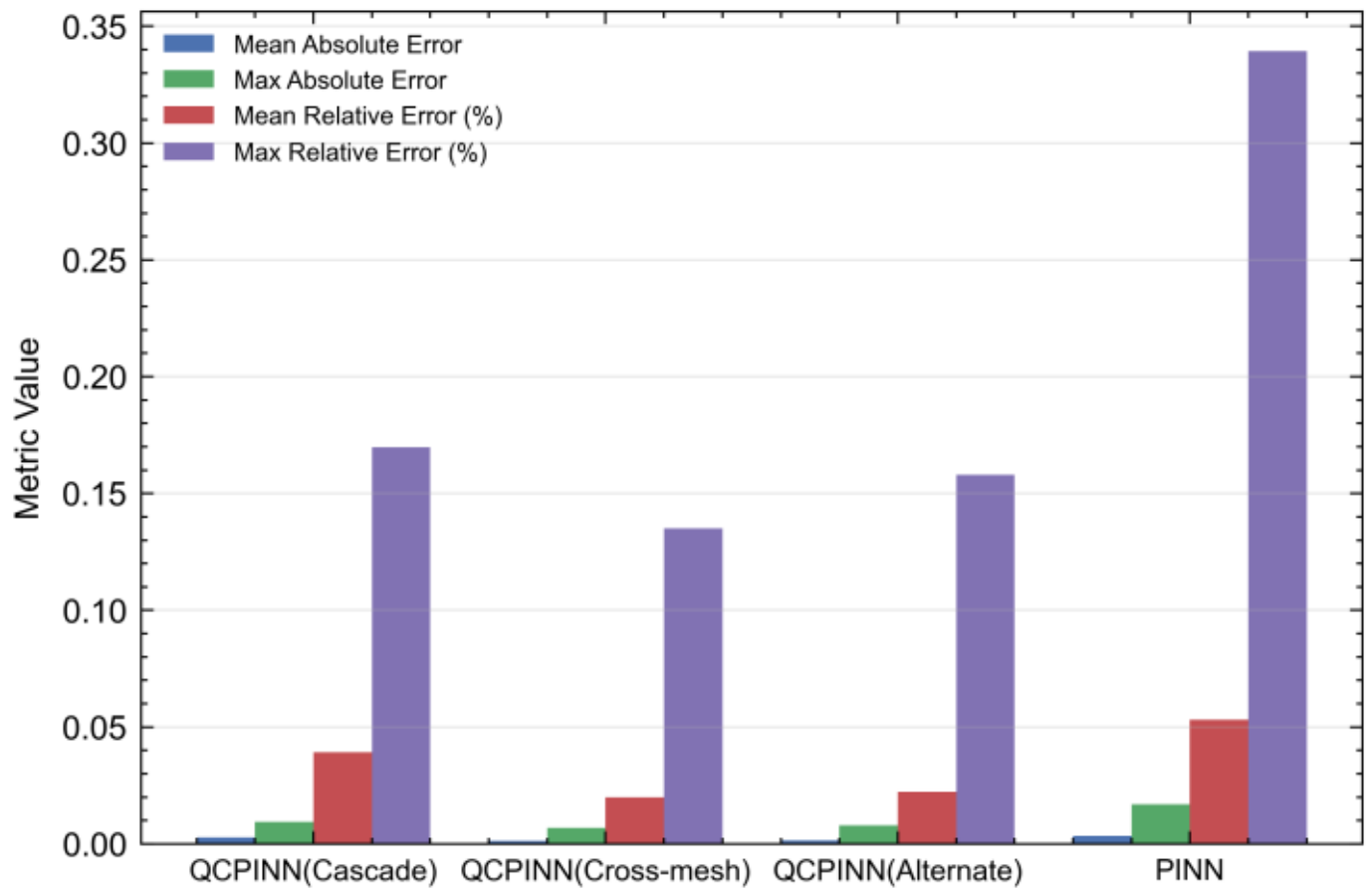

Fig. 7 Error Metrics comparison: QCPINN with three topologies vs PINN in Example 1.

Table 3 Comparison of training time of PINN and QCPINNs in Example 1.

| Method | Training Time, seconds |
|---|---|
| QCPINN (Cascade) | 246.815 |
| QCPINN (Cross-mesh) | 221.868 |
| QCPINN (Alternate) | 200.507 |
| PINN | 89.252 |

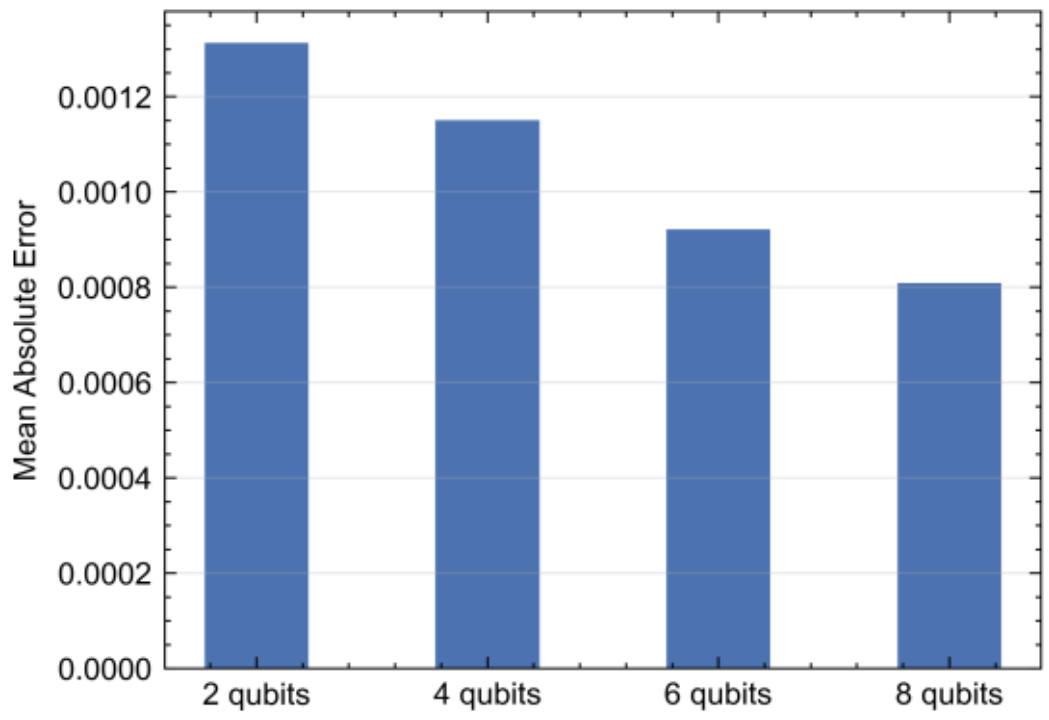

(a) Qubits sensitivity (1 Layer)

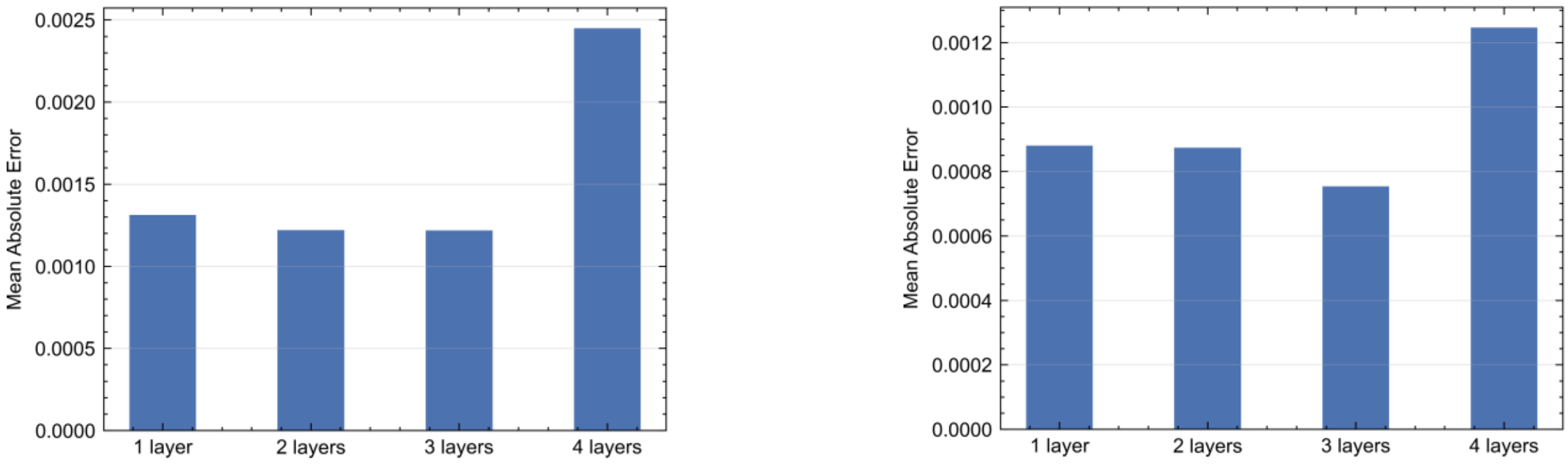

(b) Layers sensitivity (2 Qubits, batch size=64) (c) Layers sensitivity (2 Qubits, batch size=1024)

Fig. 8 Qubits and layers sensitivity of QCPINN using cross-mesh topology in Example 1.

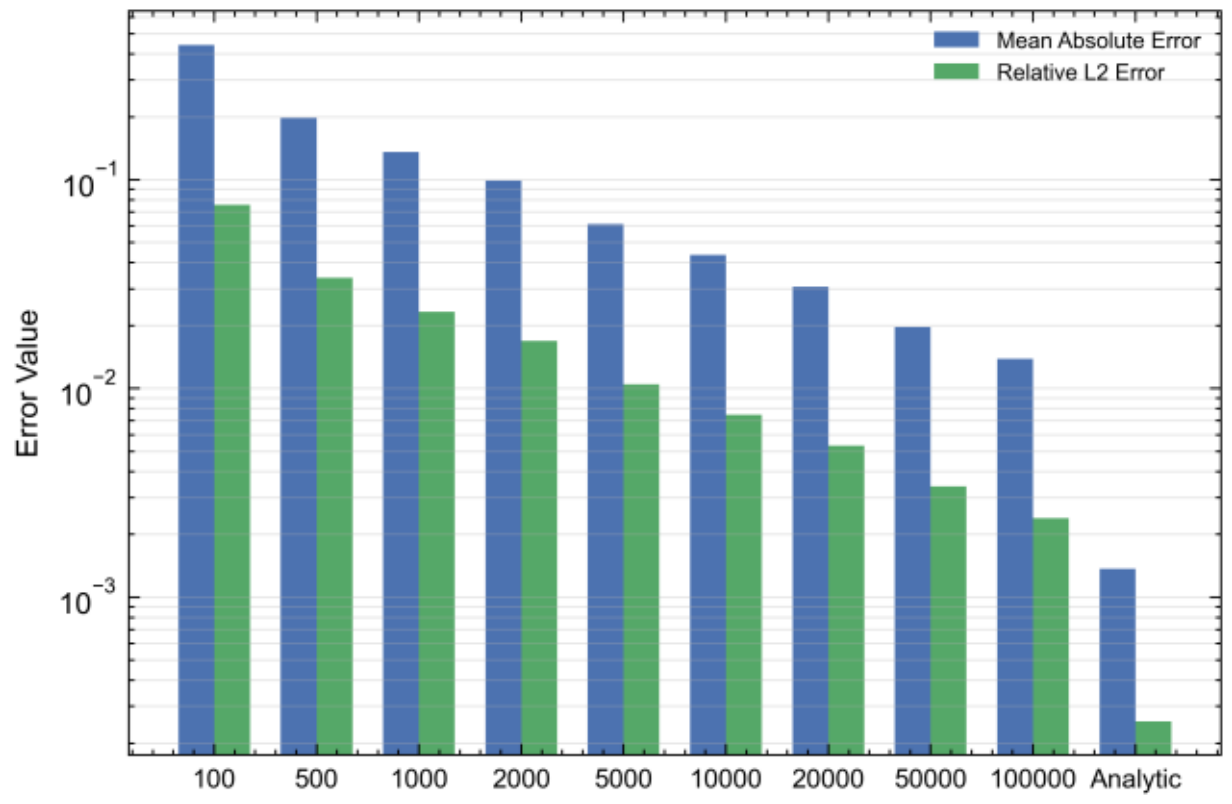

Fig. 9 Shots sensitivity of QCPINN using cross-mesh topology in Example 1.

## 4.2 Example 2

This example focuses on the two-phase flow problem described in Section 2.2. By setting $S_{wc}=0$, $S_{or}=0$, $n_o=n_w=1$, and a constant total oil-water two-phase velocity $v=1$ in the *x*-direction (the units of *x*, *t*, and $v$ are adjusted to be compatible, so they are not explicitly specified in this example), Eq. (3) can be concretized as the nonlinear Buckley-Leverett (BL) equation in Eq. (28).

$$\frac{\partial S_w}{\partial t}+v\frac{\partial f_w}{\partial x}=0,\quad 0\le x\le 1,\quad 0\le t\le 1,\quad f_w(S_w)=S_w\Big/\left(S_w+\frac{1-S_w}{M}\right) \tag{28}$$

where $M=2$ is the viscosity ratio of oil to water.

The initial water saturation in the reservoir calculation domain is $S_{wc}=0$. The left boundary is set as a constant water saturation boundary of 1 to simulate continuous water injection displacement conditions. The initial and boundary conditions of Eq. (28) are expressed as:

$$\begin{aligned} S_w(0,t)&=1,\ 0\le t\le 1,\\ S_w(x,0)&=0,\ 0\le x\le 1. \end{aligned} \tag{29}$$

The QCPINN model in this example takes spatial coordinates and time variables $(x,t)$ as input and outputs the water saturation $S_w$. To systematically evaluate the impact of quantum circuit topologies on QCPINN performance, this example uses the same three topologies (Cascade, Cross-mesh, and Alternate) as in Example 1, with the hidden-layer neuron count increased to 30. Due to the transient nature of the problem and the added nonlinearity of the water cut function, the qubit number is increased to 5 while keeping the number of quantum layers at 1. The corresponding three quantum circuit topologies are shown in Fig. 10. The PINN model used for comparative validation adopts the same structure for its first and last hidden layers as QCPINN, each containing 30 neurons. To ensure a comparable number of trainable parameters between the two models, the middle hidden layer of the PINN is set to 5 neurons. As a result, the total number of trainable parameters in the PINN model is 756, while those in the three QCPINN models are 771, 796, and 772. In addition, key training parameters of the PINN, including input and output parameters, optimizer configuration, and the number of training iterations, are kept exactly the same as those of the QCPINN to ensure the fairness of the comparative experiments. All other configuration parameters of the two models (neural network structure, optimizer settings, etc.) and the hardware environment used for running are consistent with those in Example 1.

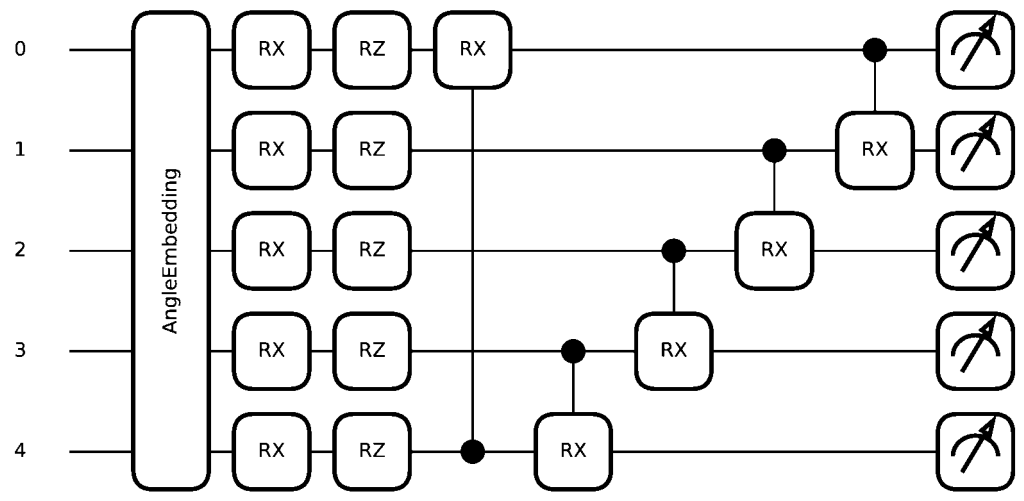

(a) Cascade

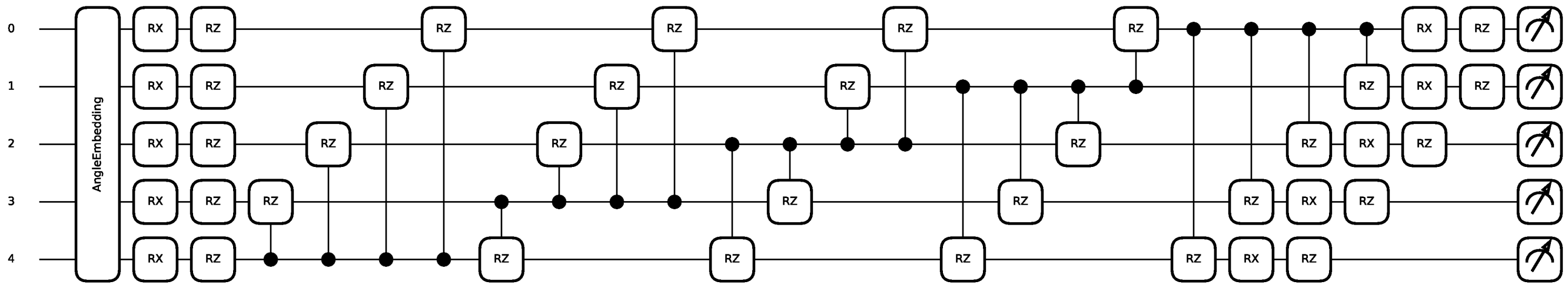

(b) Cross-mesh

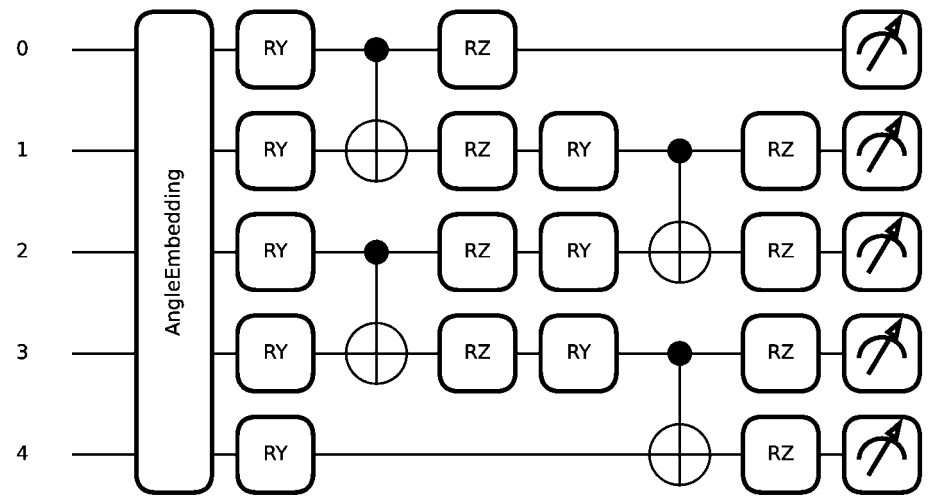

(c) Alternate

Fig. 10 Three quantum circuit topologies used in Example 2.

Fig. 11 presents a comparison of the loss function versus epochs during training for QCPINNs with three quantum topologies and the conventional PINN. Overall, all models achieve a gradual decrease in the loss function from an initial magnitude of $10^2$ to $10^{-2}$. However, the QCPINN models exhibit significantly smaller oscillations around $10^{-2}$ in the later training stages compared to the PINN. Notably, the loss function curve of the PINN displays intense and irregular fluctuations, reflecting its training instability when dealing with this strongly nonlinear transient problem. In contrast, although the three QCPINN models also exhibit some oscillations, their overall convergence trends are considerably more stable. Specifically, the QCPINN with the Cross-mesh topology shows the fastest loss decay in the early training stages; the QCPINN with the Cascade topology experiences relatively pronounced fluctuations in the later stages, posing certain challenges to the convergence of its loss function during training; the QCPINN with the Alternate topology demonstrates the smallest oscillation amplitude in its loss function curve, exhibiting the best training stability. Ultimately, the loss levels of the three QCPINN models converge to a consistent magnitude of $10^{-2}$. While the final loss of the PINN also approaches this magnitude, its pervasive fluctuating behavior throughout training indicates inferior convergence compared to the QCPINNs.

We compute the reference solution using the upwind FDM with forward-Euler scheme. The first-order upwind scheme is adopted for the convective term, with 200 uniform grid points in the 1D spatial domain and 1000 uniform time steps. Fig. 12 compares the prediction results of the three QCPINN topologies at different time instants ($t$=0.1, 0.3, 0.5, 0.7, 0.9) with the reference solution. The results demonstrate that all three topologies can accurately capture the steep water-flooding front and its migration behavior, with the overall morphology being basically consistent with the reference solution. Specifically, the Cascade topology shows slight deviations in the front and the water-unswept regions, but the overall curve trend remains consistent with the reference solution. The Cross-mesh topology is highly consistent with the reference solution in all regions at each time step except for a noticeable deviation from the reference solution at the front. The Alternate topology only exhibits minor deviations in the water-unswept regions, while fitting most closely with the reference solution in other regions, almost overlapping completely.

Fig. 13 presents the relative L2 errors between the saturation predictions of QCPINN using three quantum-circuit topologies, PINN and the reference solution over the time interval of 0.1-0.9. As a quantitative evaluation metric, the L2 error is defined as follows:

$$L_2\left(t_k\right)=\sqrt{\frac{1}{N}\sum_{i=1}^{N}\left(S_{pred}\left(x_i,t_k\right)-S_{ref}\left(x_i,t_k\right)\right)^2}, \tag{30}$$

where $N$ denotes the number of spatial sampling points, $t_k$ represents a specific time point, $S_{pred}$ and $S_{ref}$ stand for the water saturation predicted by QCPINN, PINN and the reference solution, respectively.

We can see from Fig. 13 that the L2 error of the PINN remains consistently higher than those of the QCPINN models using three topologies across all time steps except $t$= 0.1, with a maximum error approaching 0.0225. The errors of the three QCPINN models remain at a low magnitude, indicating the QCPINN exhibits excellent accuracy. Specifically, in the early stage $t$= 0.1, the Cross-mesh topology achieves the lowest error (approximately 0.008), followed by the Cascade topology, while the Alternate topology yields the highest error. With the time increasing, the accuracy performance of each topology undergoes significant changes. The error of the Cross-mesh topology gradually increases, whereas the error of the Alternate topology decreases markedly and remains consistently at a low level (around 0.006) for all time steps from $t$= 0.4 to $t$= 0.9, demonstrating significantly superior accuracy compared to the Cascade and Cross-mesh topologies. The Cascade topology exhibits relatively high errors across all time steps and fails to demonstrate obvious advantages at any stage. Overall, for the 1D nonlinear Buckley-Leverett (BL) equation describing waterflooding, compared with PINN, the QCPINN models constructed with the Cascade, Cross-mesh, and Alternate topologies all demonstrate better capabilities in capturing physical characteristics and higher prediction accuracy. Among them, the Alternate topology stands out in terms of training convergence and computational accuracy, showing relatively superior performance. Table 4 presents a comparison of training times between classical PINN and QCPINNs in Example 2. Consistent with the observations in Example 1, the classical PINN achieves the shortest training time of 77.618s on the GPU-accelerated platform. The QCPINN models, however, require longer training times, ranging from 239.862s to 278.758s, with the Cascade topology exhibiting the highest computational efficiency among the quantum variants. The increased runtime of QCPINNs is attributed to the computational overhead of simulating quantum superposition and entanglement on classical hardware, despite the significant acceleration provided by GPU. These results further confirm that QCPINNs deliver superior predictive accuracy for two-phase flow in heterogeneous reservoirs at the cost of moderate additional training time on classical GPUs.

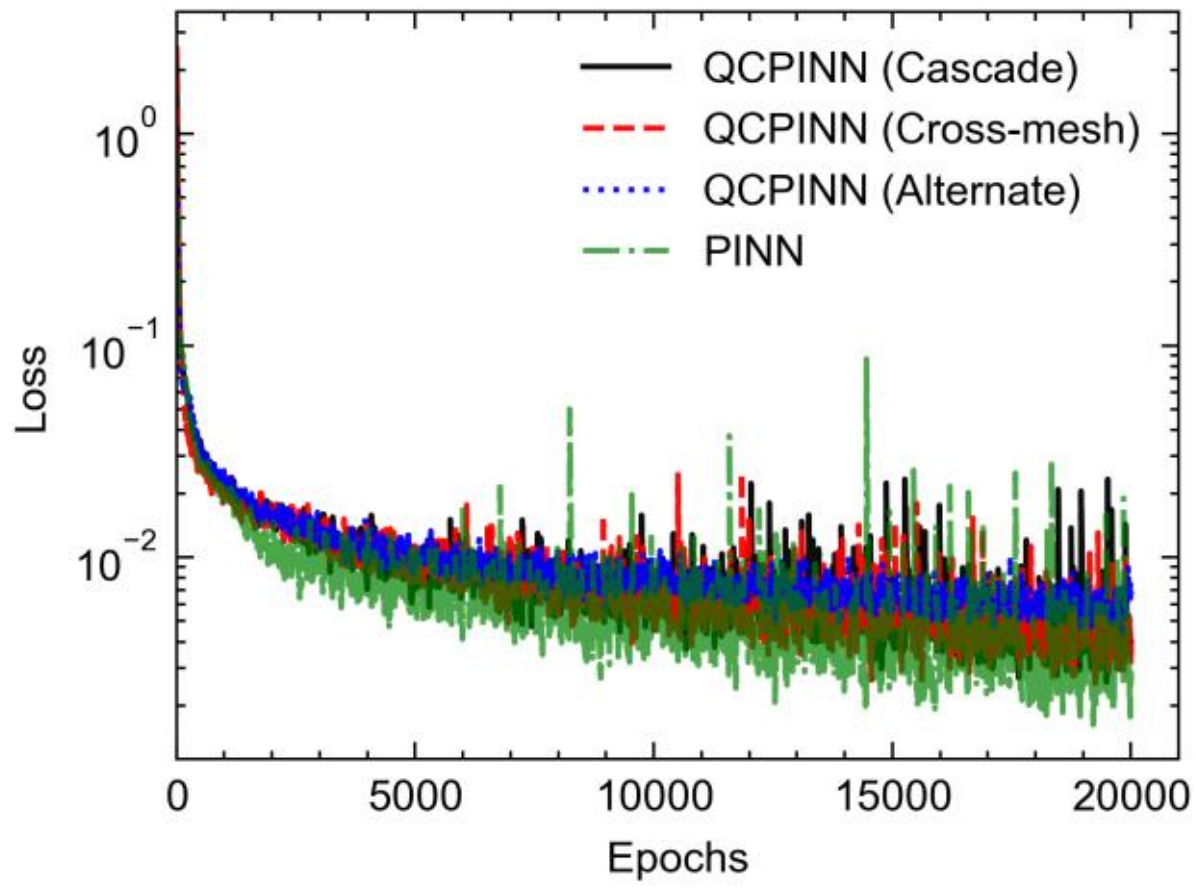

Fig. 11 Comparison of loss function versus epochs of QCPINN with three quantum circuit topologies and PINN in Example 2.

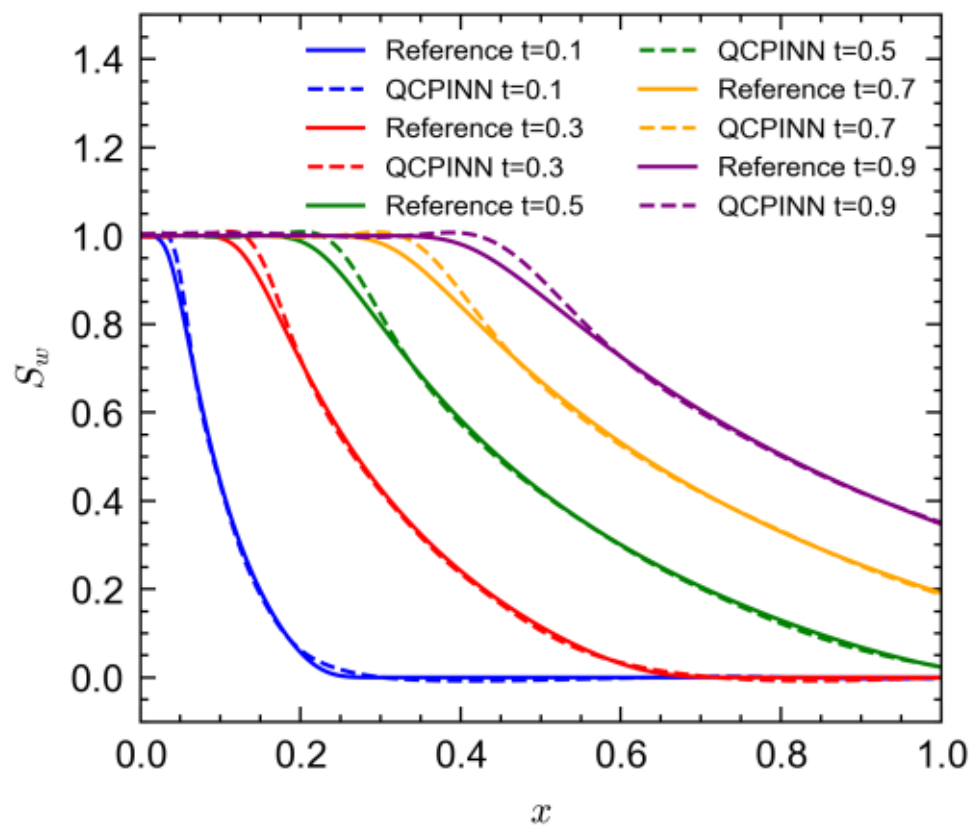

(a) PINN solution

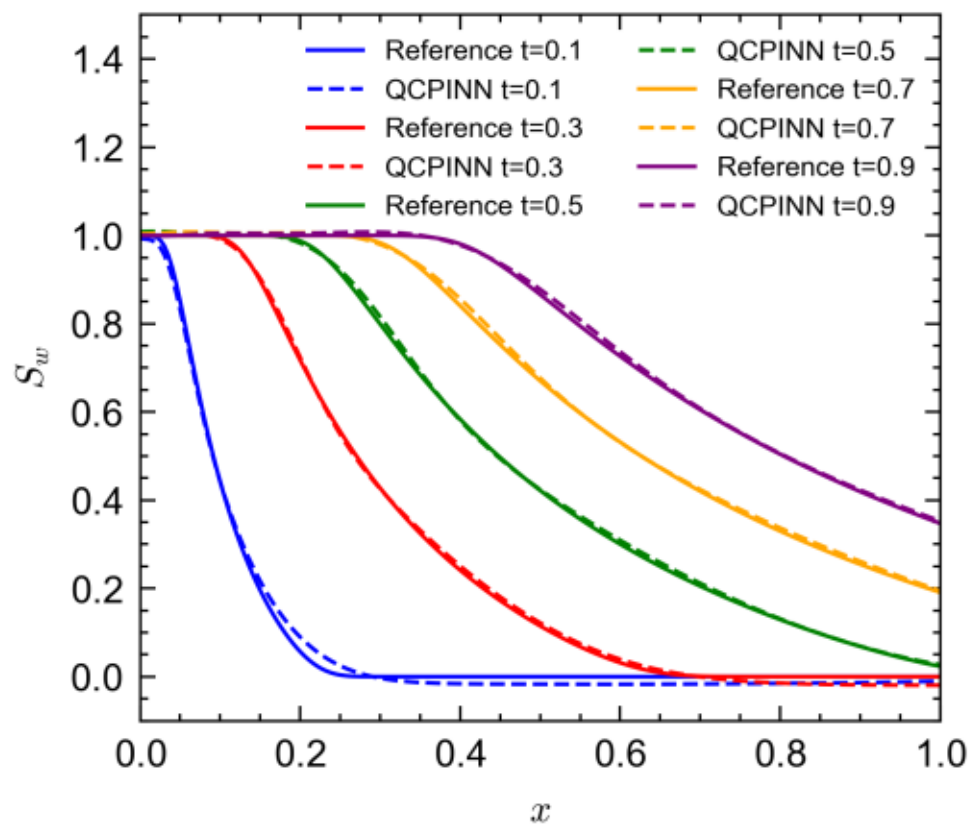

(b) QCPINN solution (Cascade)

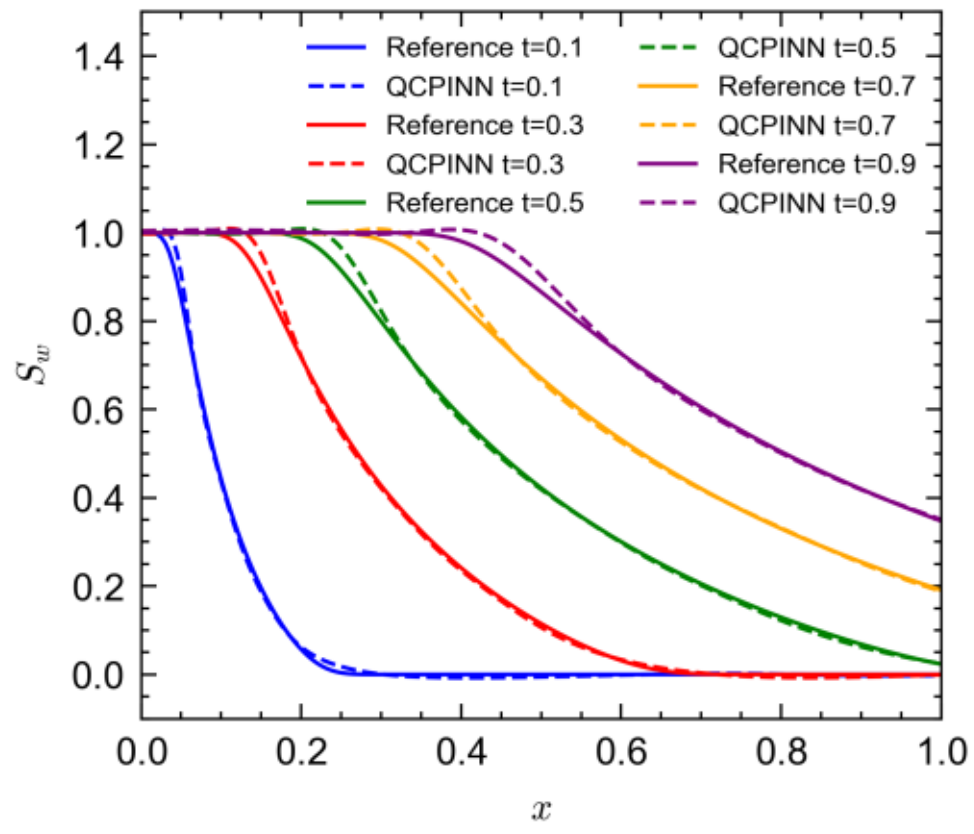

(c) QCPINN solution (Cross-mesh)

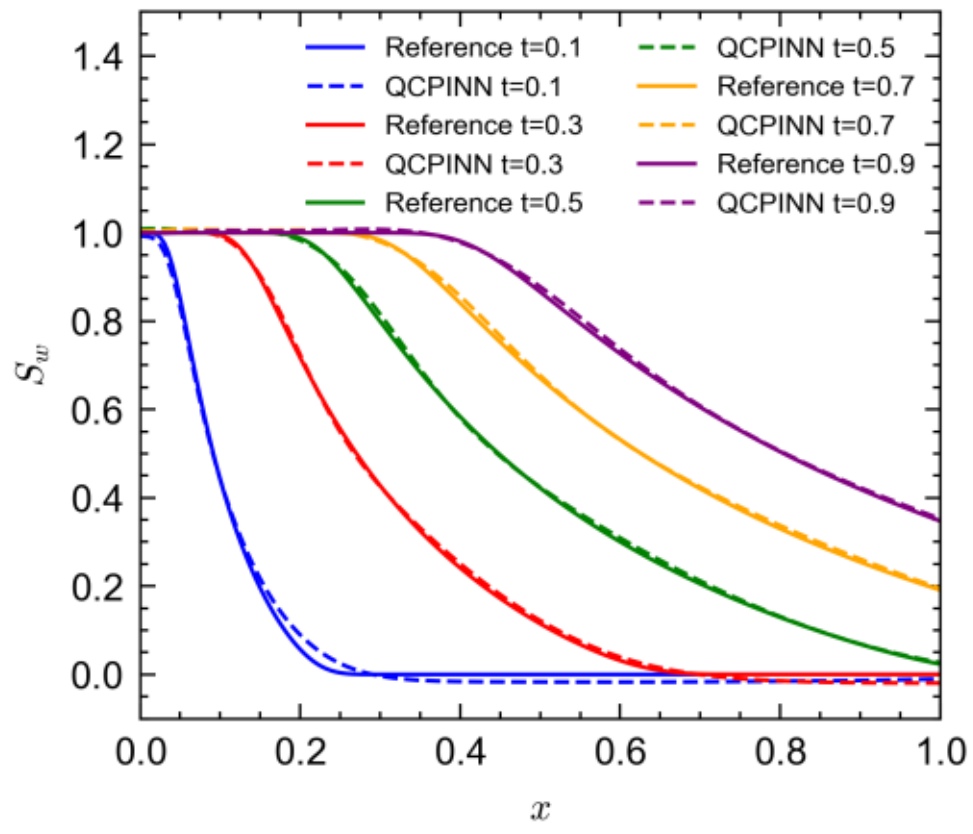

(d) QCPINN solution (Alternate)

Fig. 12 Comparison of saturation distributions: QCPINN predictions using three topologies, PINN vs reference solution at different times.

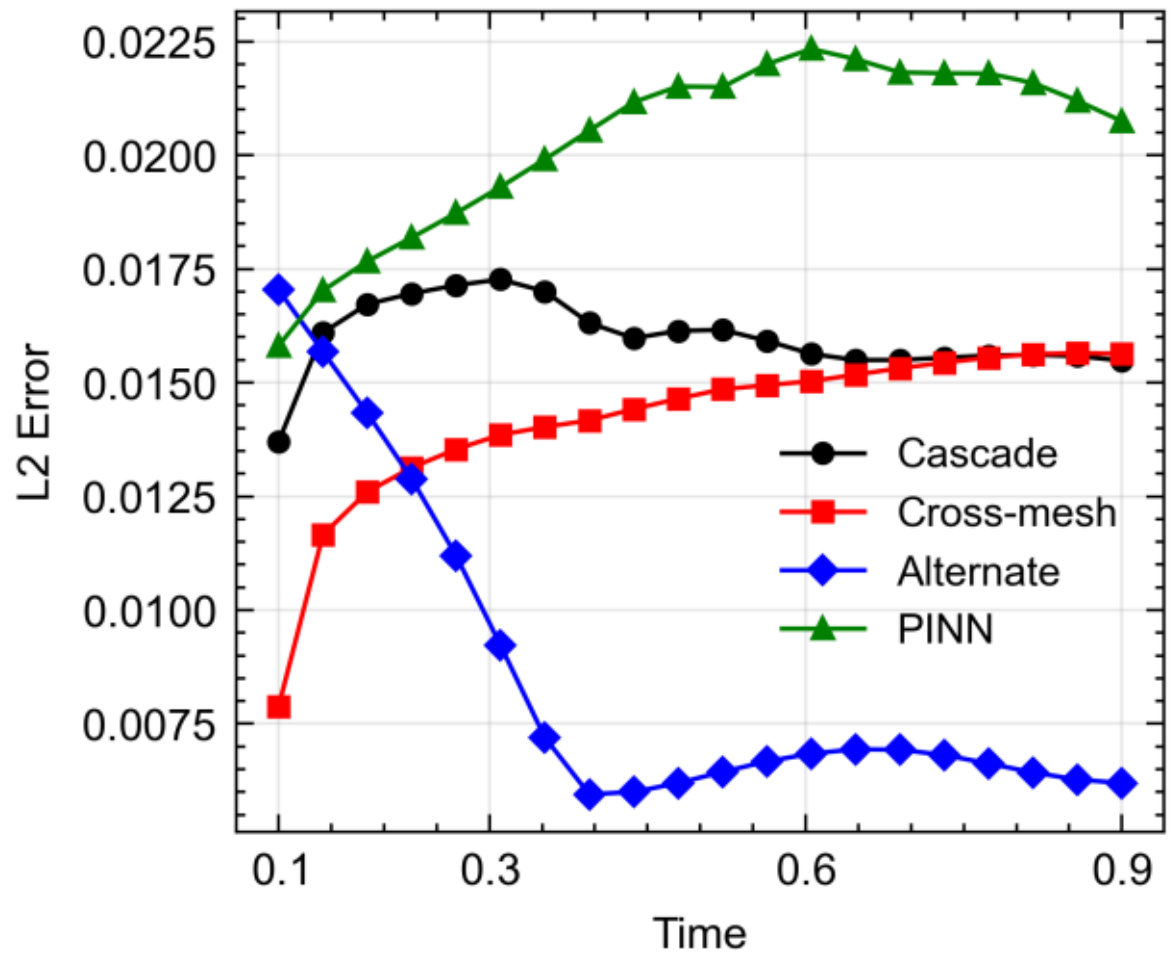

Fig. 13 Relative L2 error comparison of three topologies in Example 2.

Table 4 Comparison of training time of PINN and QCPINNs in Example 2.

| Method | Training Time, seconds |
|---|---|
| QCPINN (Cascade) | 249.178 |
| QCPINN (Cross-mesh) | 278.758 |
| QCPINN (Alternate) | 239.862 |
| PINN | 77.618 |

### 4.3 Example 3

This example focuses on the component transport problem considering adsorption as discussed in Section 2.3. As illustrated in Fig. 14, a rectangular reservoir computational domain Ω= [0, 100]m×[0, 100]m is still adopted. In Eq. (6), suppose $\phi = 0.3$, $\rho_b = 2000$ kg/m$^3$, $K_d = 1.0\times10^{-4}$ m$^3$/kg, $D = 0.5$ m$^2$/s, and $\mathbf{v} = (0.5, 0)$ m/s. At the initial moment, the component concentration throughout the reservoir is 0 kg/m$^3$. A Dirichlet boundary condition is applied to the left boundary, where the solute concentration is maintained at 1 kg/m$^3$; the right boundary has a zero diffusion flux but no restriction on advective flux. The upper and lower boundaries are set as no-flux closed boundaries. This example illustrates solute transport driven by convection, dispersion and linear adsorption. Linear adsorption slows solute migration, making its effective velocity lower than seepage velocity. Initially, solutes diffuse from the high-concentration left boundary to low-concentration zones, generating an adsorption-retarded concentration front moving along the x-axis. Dispersion widens the front as it advances. The coupled process is dominated by convection, with transport delayed by adsorption and spread by dispersion.

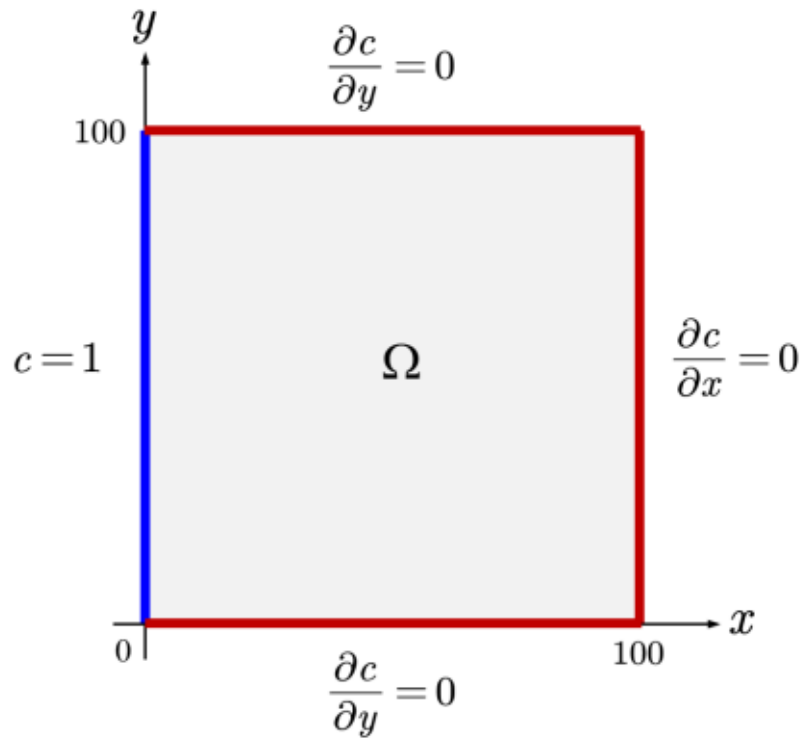

Fig. 14 Schematic of the computational domain and boundary conditions for Example 3.

The QCPINN model takes spatiotemporal coordinates ($x$, $y$, $t$) as input and outputs the target concentration $c$. Since this example involves multi-physical processes, to accurately characterize the component migration behavior, 4 qubits are used in this case, and three quantum topologies (Cascade, Cross-mesh, and Alternate) are still adopted as shown in Fig. 15. The remaining parameters and training configurations of the model are the same as those in Example 2. To demonstrate the performance advantages of QCPINN, a PINN model is also introduced as a comparator in this example. The input layer, output layer, and first/last hidden layers of this PINN are structurally identical to those of QCPINN. By setting the number of neurons in the middle-hidden layer to 4, the total number of trainable parameters in the PINN reaches 425, which is comparable to the 437, 453, and 441 trainable parameters of QCPINN, ensuring fairness in the comparison. In addition, key training parameters of the PINN, such as optimizer configuration and the number of training iterations, are kept consistent with those of QCPINN to eliminate the influence of differences in training conditions on performance evaluation.

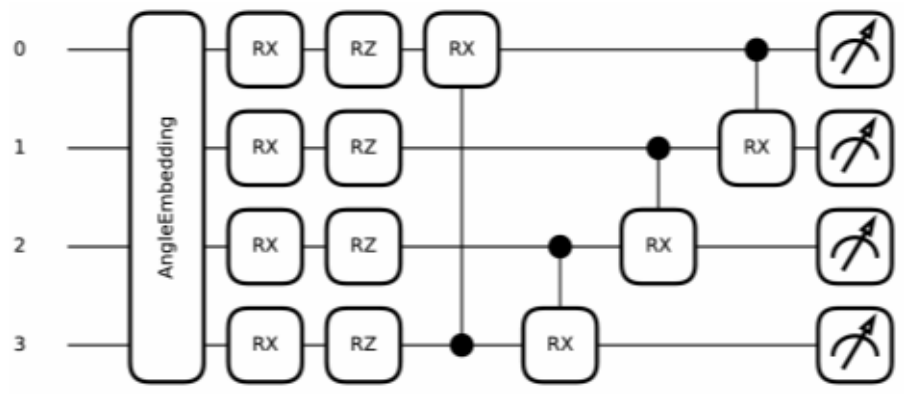

(a) Cascade

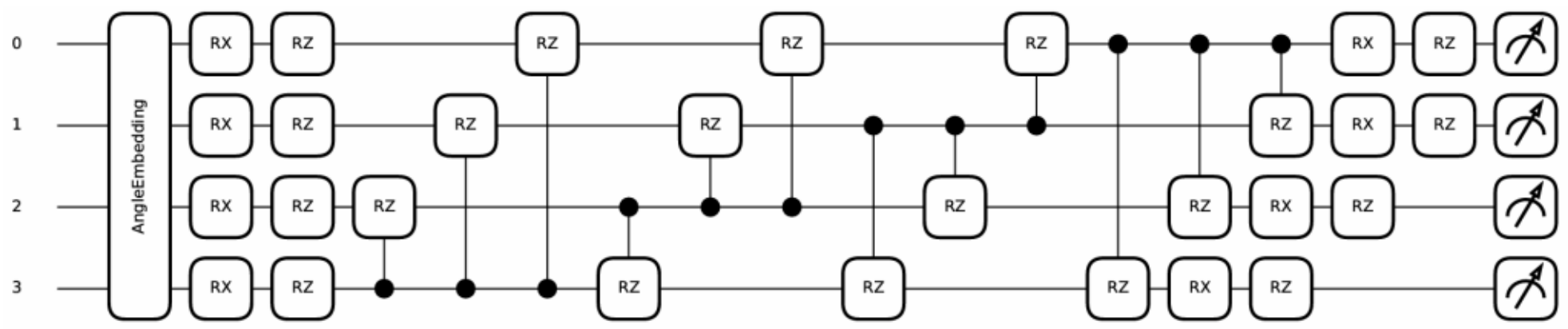

(b) Cross-mesh

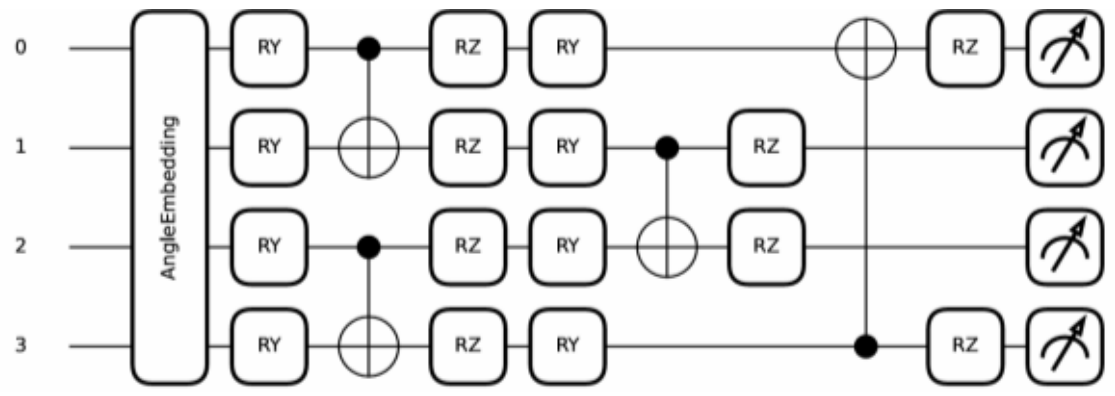

(c) Alternate

Fig. 15 Three quantum circuit topologies used in Example 3.

Fig. 16 shows the convergence histories of the loss function for the QCPINN models with three quantum circuit topologies, together with the traditional PINN. Overall, all four models exhibit effective convergence, with the loss decreasing from the initial order of $10^0$ to approximately $10^{-2}$. A clear difference appears after about 1000 training epochs: the QCPINN models continue to reduce the loss more efficiently, whereas the PINN shows a relatively slower decrease and remains at a higher loss level during the later training stage. Among the three QCPINN topologies, the Alternate topology achieves the most favorable convergence behavior, showing a faster loss reduction after the early training stage and reaching the lowest final loss. The Cascade topology also maintains a steady decreasing trend, while the Cross-mesh topology exhibits relatively noticeable fluctuations during training.

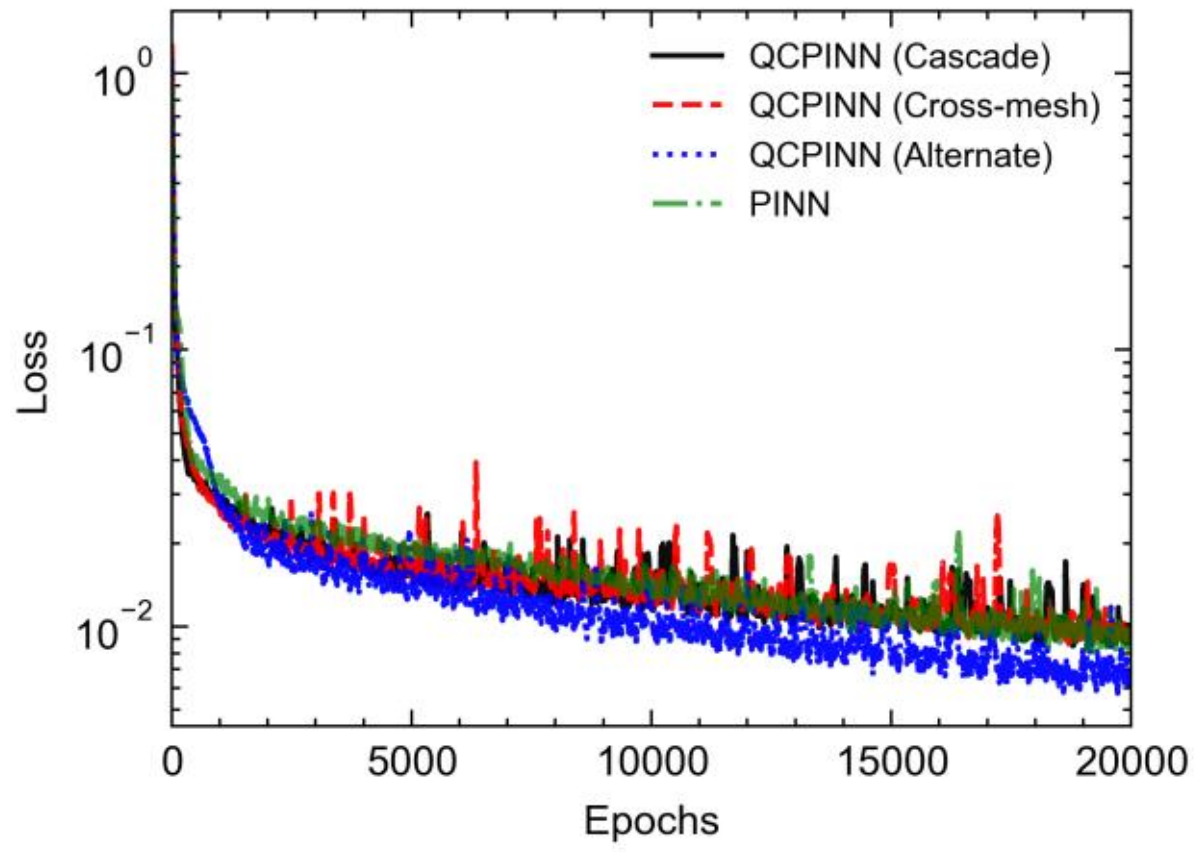

Fig. 16 Comparison of loss function versus epochs of QCPINN with three quantum circuit topologies and PINN in Example 3.

We use the upwind FDM to compute the reference solution for this example, where the spatial domain employs 200×200 =40,000 grid points, and the temporal domain uses the same 1000 time steps as Example 2. Fig. 17 shows the comparisons of concentration distributions between the reference solution and the predictions of the QCPINNs and the PINN at $t$=0.8, and further presents the corresponding absolute error distributions. The main errors of all models are concentrated in the component displacement front, where the concentration varies sharply. This is because the large concentration gradient in this region significantly increases the approximation difficulty for neural networks. A detailed comparison shows that the absolute error of the PINN in the front region is markedly higher than that of the three QCPINN models, with its peak error approaching 0.16, reflecting its insufficient accuracy in capturing regions with high gradients. In contrast, the three QCPINN models exhibit overall lower error levels. Among them, the QCPINN with the Alternate topology shows the most uniform error distribution and the smallest peak error in the component displacement front, demonstrating the best local approximation capability.

Figure 18 assesses transient solute transport predictions at monitoring point (50, 50) for the PINN and three QCPINN topologies (Alternate, Cascade, Cross-mesh). Figure 18(a) shows all methods reproduce the typical S-shaped concentration curve characterizing concentration front breakthrough and stabilization. Discrepancies mainly emerge during the steep concentration rise near front arrival; PINN deviates noticeably from the reference solution, while Alternate QCPINN matches the benchmark best throughout the entire simulation period. The L2 concentration errors in Fig. 18(b) verify this trend: Alternate ($1.6137\times10^{-2}$) outperforms Cascade ($3.2797\times10^{-2}$), Cross-mesh ($3.6374\times10^{-2}$) and PINN ($4.0437\times10^{-2}$), cutting concentration prediction error by 60.1% relative to PINN. Temporal concentration gradients in Fig. 18(c) serve as a stricter evaluation metric for front propagation velocity and sharp transient shifts. Though all models capture the primary gradient peak at breakthrough, PINN generates wider, lagged gradient responses with substantial mismatches at the peak and decay phases. All QCPINNs better replicate reference gradient dynamics, and Alternate topology delivers the most consistent peak morphology and decay trend. The gradient $L_2$ errors in Fig. 18(d) corroborate this result: Alternate reaches $1.3225\times10^{-1}$, below Cascade ($2.3936\times10^{-1}$), Cross-mesh ($2.6290\times10^{-1}$) and PINN ($3.1707\times10^{-1}$), reducing gradient error by 58.3% versus PINN.

Overall, QCPINN frameworks outperform conventional PINN in local transient forecasting, and the Alternate topology attains optimal accuracy for both concentration and concentration-gradient evolution.

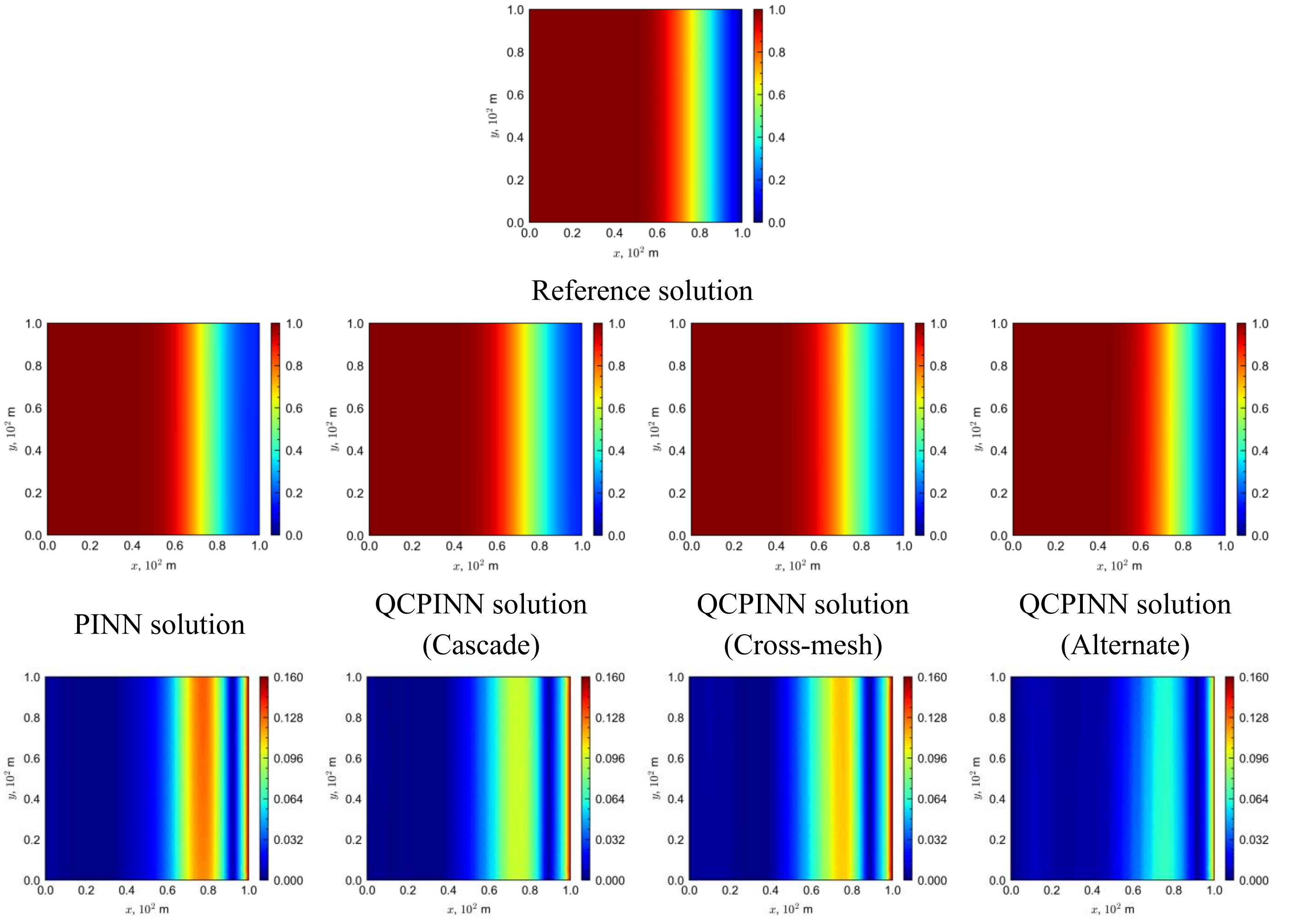

Reference solution

PINN solution

QCPINN solution (Cascade)

QCPINN solution (Cross-mesh)

QCPINN solution (Alternate)

Error distribution for PINN

Error distribution for QCPINN (Cascade)

Error distribution of QCPINN (Cross-mesh)

Error distribution of QCPINN (Alternate)

Fig. 17 Comparison of concentration distributions and absolute error distributions of QCPINN predictions using three topologies, PINN prediction vs reference solution at $t$=0.8 in Example 3.

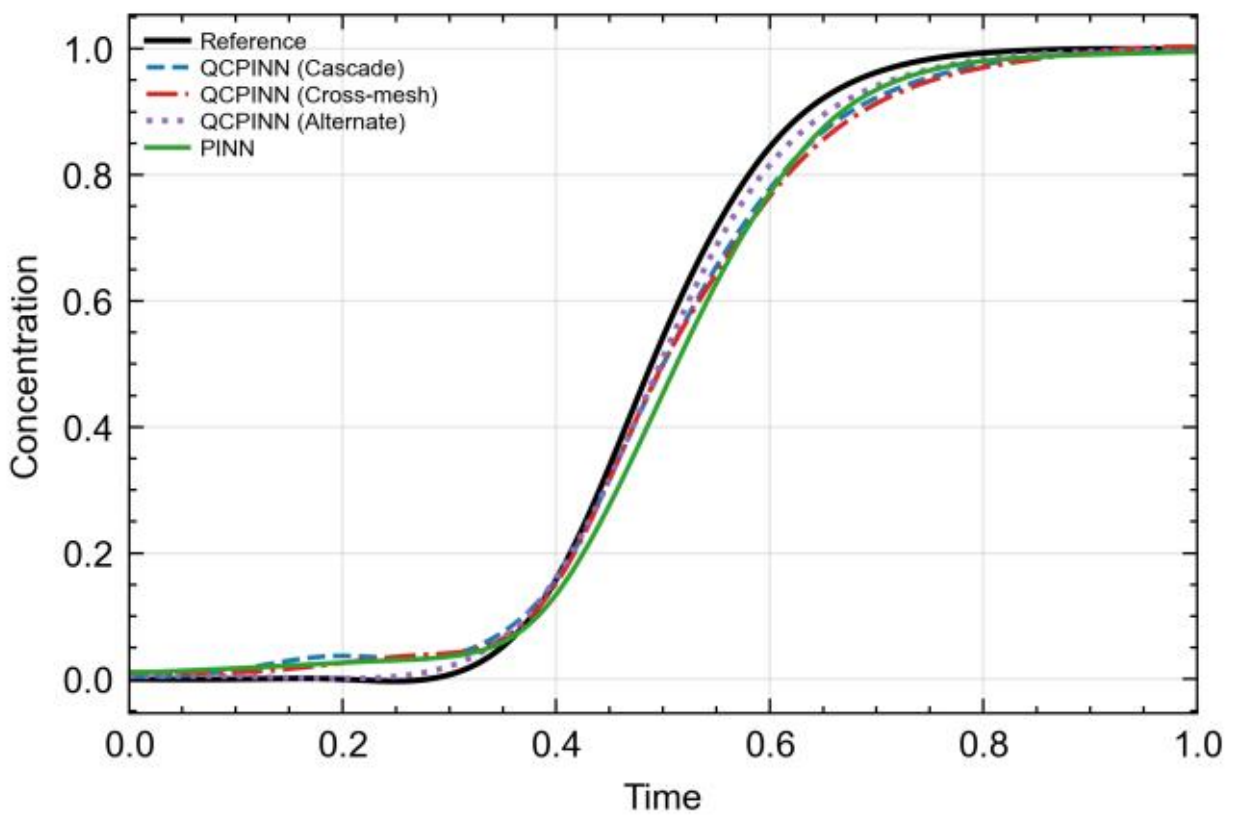

(a) Concentration evolution

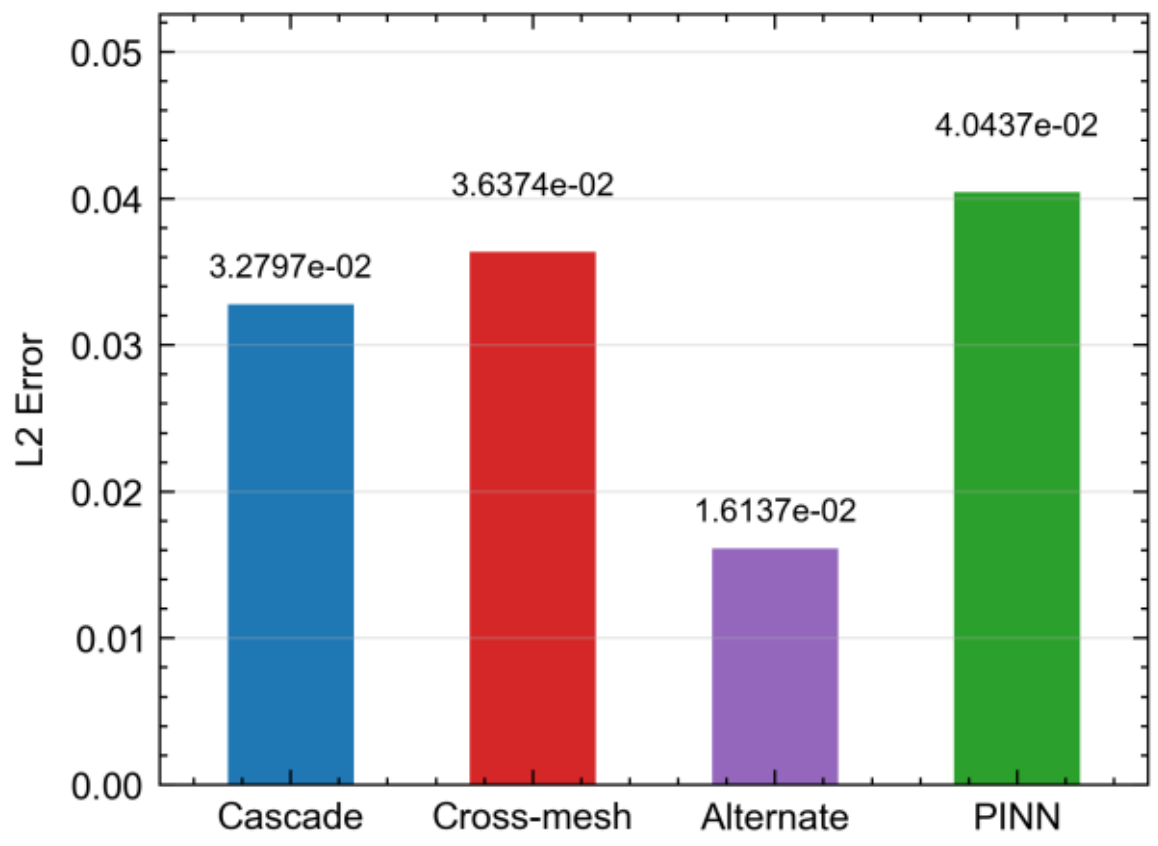

(b) L2 error of Concentration

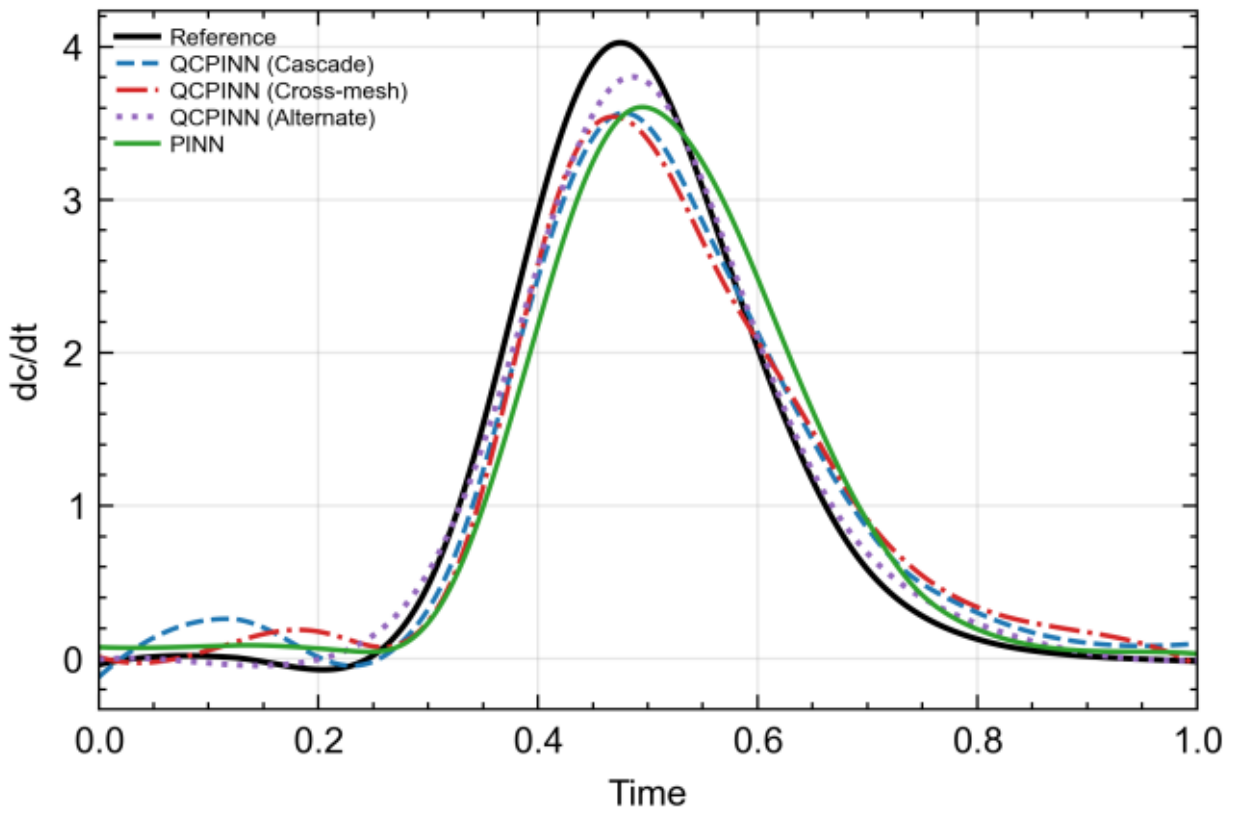

(c) Concentration-gradient evolution

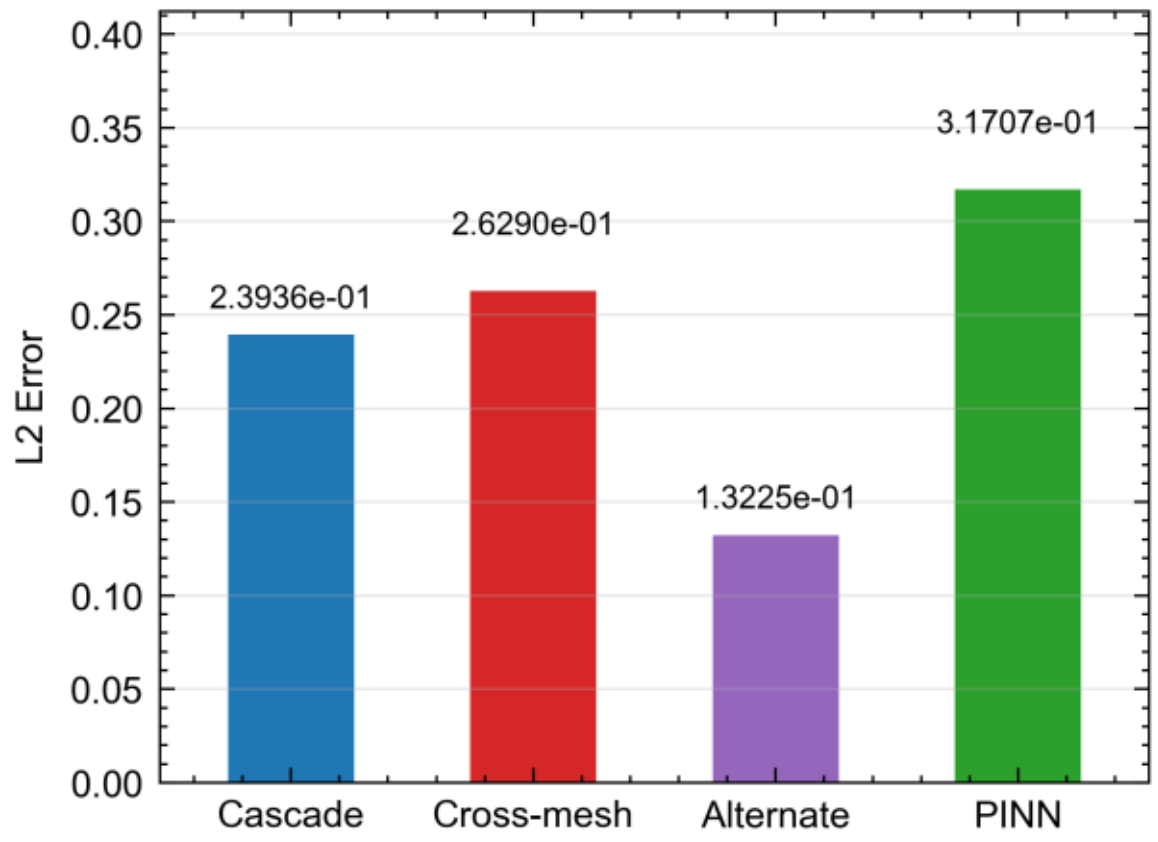

(d) L2 error of Concentration-gradient

Fig. 18 Concentration and concentration-gradient prediction at the fixed point (50, 50) in Example 3.

### 4.4 Example 4

This example focuses on the two-phase oil-water seepage problem in heterogeneous reservoirs where pressure and water saturation need to be calculated synchronously, which is different from solving the Buckley-Leverett equation that only calculates water saturation with compressibility neglected in Example 2. Compared with Example 2, two equations in Eq. (31) need to be handled synchronously in this example, which are the mass conservation equations for the water phase and oil phase, respectively. The independent variables are x, y, and t, and the dependent variables include both pressure and water saturation. In addition, the exponential strongly heterogeneous permeability distribution in Eq. (32) is introduced, which represents the most common and classical two-phase flow scenario in reservoir engineering. The only difference is that the singular source-sink term caused by injection-production wells, which is common in reservoir engineering, is not added in this example. This is because numerous previous studies (Karniadakis et al., 2021; Zhang et al., 2022; Han et al., 2023; Rao et al., 2025b) have shown that conventional PINN methods that only rely on automatic differentiation to construct equation residual terms without additional processing cannot effectively handle singular source-sink terms. At present, some improved PINNs are generally adopted to deal with such singular source-sink terms. For instance, the PINN based on domain decomposition (PINN-DD) that divides the reservoir domain into well-containing and well-free sub-domains (Han et al., 2023; Kharazmi et al., 2021), or PINNs based on numerical method formats with integral processing, such as the boundary-integral neural network (BINN) based on the boundary element method (Rao et al., 2025c), the variational physics-informed neural network (VPINN) based on the finite element method (Kharazmi et al., 2019), the enriched-PINN (E-PINN) based on the finite volume method (Yan et al., 2024), or some regularization methods that approximate the singular Dirac function with Gaussian functions or truncated polynomial functions. Among the above processing methods, only the regularization method is directly compatible with the conventional PINN framework, but its computational performance is generally poor. Therefore, the regularization method is not used to add the singular source-sink term caused by injection-production wells in this example, because the large error inherent in the regularization method will significantly weaken the illustration of the computational performance advantages of the QCPINN proposed in this work. Other processing methods for singular source-sink terms require underlying modifications to the conventional PINN framework, and such modifications will completely alter the method architecture in Section 3 of this paper. Thus, adding domain decomposition, boundary integral, weak form or finite volume processing for singular source-sink terms on the basis of this work to form the corresponding BIQCPINN, QCPINN-DD, VQCPINN or E-QCPINN is fully worthy of a complete new research. Hence, this example focuses on the classical two-phase oil-water flow simulation in heterogeneous reservoirs without source-sink terms to fully illustrate the performance advantages of the quantum computing-coupled QCPINN over the traditional PINN. It can be fully expected that since PINN-based methods such as BINN, PINN-DD, VPINN and E-PINN can achieve excellent computational performance for seepage problems involving singular source-sink terms, BIQCPINN, QCPINN-DD, VQCPINN, and E-QCPINN which are based on the more superior QCPINN, will also effectively handle seepage problems with singular source-sink terms.

$$
\begin{aligned}
&\phi\frac{\partial S_w}{\partial t}+\nabla\cdot\left(\frac{k(x,y)k_{rw}(S_w)}{\mu_w}\nabla p\right)=0,\\
&\phi\frac{\partial S_o}{\partial t}+\nabla\cdot\left(\frac{k(x,y)k_{ro}(S_w)}{\mu_o}\nabla p\right)=0,
\end{aligned}
\tag{31}
$$

$$
k(x,y)=0.01\times e^{\frac{x^2+y^2}{5000}}D, \tag{32}
$$

where $S_w$ is the water saturation and $p$ is the oil phase pressure; $k(x,y)$ is the absolute permeability, which is a function of spatial coordinates. $k_{rw}(S_w)$ and $k_{ro}(S_w)$ are the relative permeabilities of the water phase and oil phase, respectively, which are nonlinear functions of saturation and described by the Corey model in Eq. (5). For the convenience of comparative analysis, $n_w=1$, $n_o=1$, $\mu_w=1cp$, and $\mu_o=1cp$ are taken in this example.

As shown in Fig. 19(a), the computational domain of the two-dimensional reservoir in this example is Ω= [0, 100]m×[0, 100]m. For the initial conditions, $S_w(x,\ 0)=S_{wc}$ and $S_{wc}=0$ are adopted in this example. The upper and lower boundaries are closed boundaries satisfying the no-flow condition with $\nabla S_w(x,t)\cdot\mathbf{n}=0$ and $\nabla p(x,t)\cdot\mathbf{n}=0$. The left boundary is a constant-pressure water injection boundary with a pressure of 10 MPa and $S_w(0,\ t)=1-S_{or}$, where $S_{or}=0$ are taken in this example, and the pressure of the right boundary is 5 MPa. As shown in Fig. 19(b), the exponential permeability distribution in Eq. (32) is adopted in this example, which generates a drastic change of more than an order of magnitude in the computational domain to fully verify the computational performance of QCPINN.

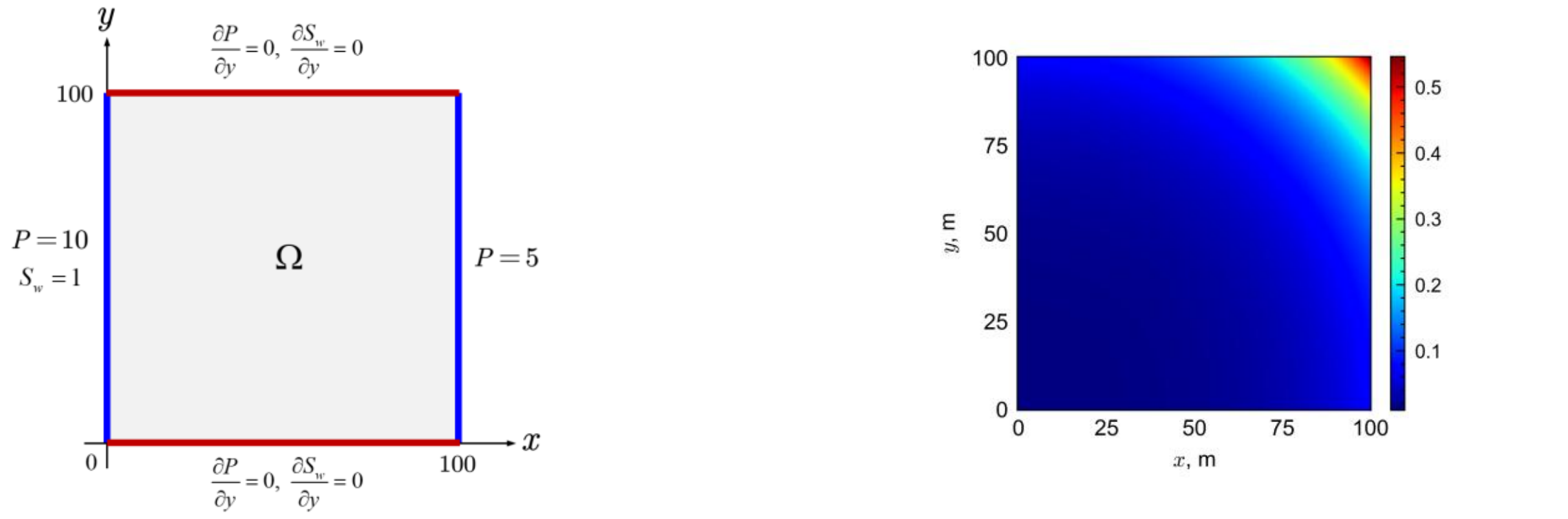

(a) Reservoir domain and boundary conditions (b) Heterogeneous permeability distribution

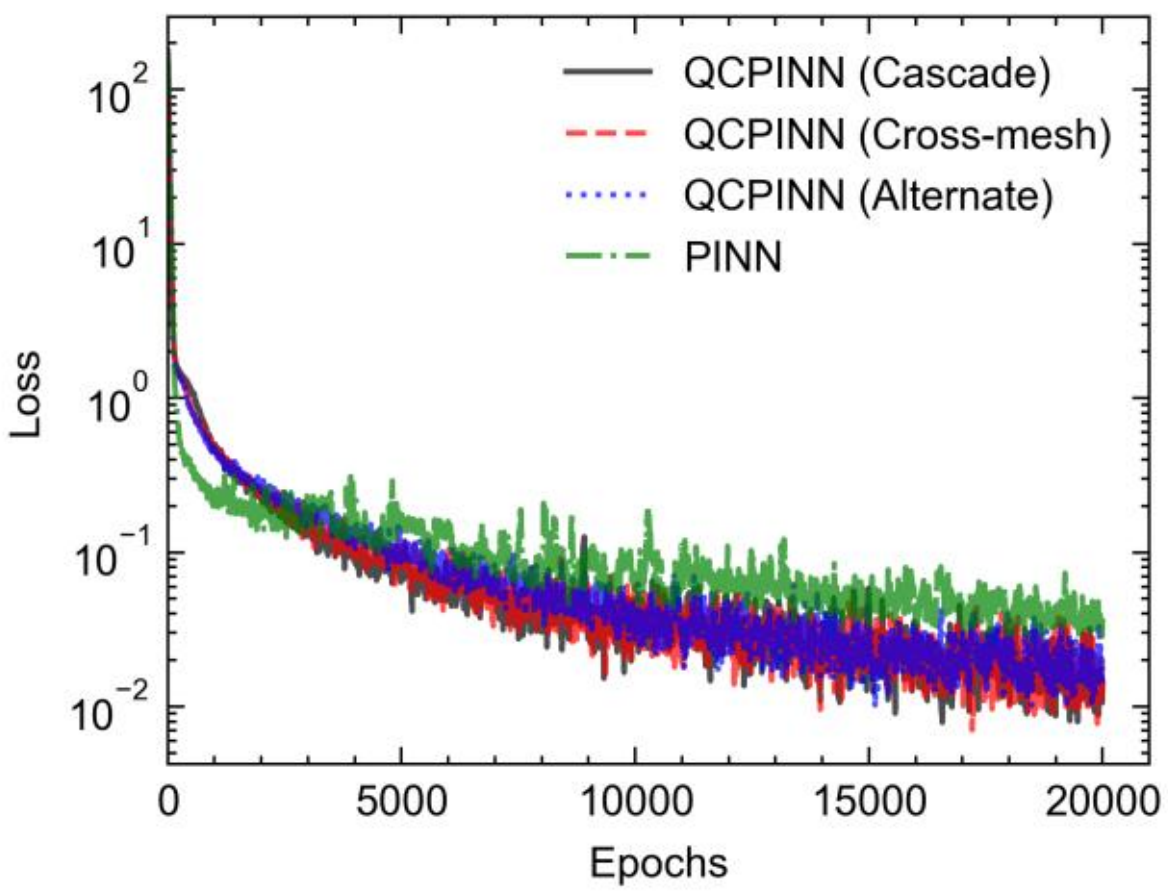

(c) Training loss over epochs

Fig. 19 Schematic of boundary conditions for the reservoir domain, heterogeneous permeability distribution and training loss over epochs of Example 4.

Due to the significant differences in physical magnitudes and dynamic characteristics between pressure and saturation, to avoid the gradient imbalance problem caused by inconsistent output scales of a single network, two independent QCPINN sub-networks are constructed in this example to approximate pressure and saturation, respectively. The configuration of the quantum core of these networks (including the number of qubits, layers and quantum circuit topology) is the same as that in Example 3. The losses of the PDE residual, boundary conditions and initial conditions for Eq. (31) constituted by the two sub-networks are calculated separately and then summed, and the optimization is performed by synchronously updating the parameters of the two networks. The parameter settings such as optimizer, learning rate strategy and batch size are consistent with those in Example 3.

To quantitatively evaluate the performance improvement of QCPINN over PINN in this example, a classical PINN model for comparison is trained simultaneously, as in the previous examples. The structure of the input layer, output layer and the first and last hidden layers of the PINN model is exactly the same as that of QCPINN, and the number of neurons in the middle hidden layers is 6, the same as the number of qubits in QCPINN, to ensure a comparable scale of trainable parameters. The PINN also uses two independent sub-networks to output pressure and saturation, respectively, to ensure the fairness of comparison.

Fig. 19(c) shows the loss functions versus epochs of the QCPINN models with three quantum circuit topologies and the traditional PINN model during the training process. Overall, the loss values of all models gradually decrease from the initial order of 100 to about the order of $10^{-2}$. However, the loss function value of the traditional PINN remains at a relatively higher level than those of the QCPINNs after 3000 epochs, and its loss function convergence ability is significantly weaker than that of the three QCPINN models. The convergence of the three QCPINN models is relatively consistent with similar final loss function values, indicating that QCPINN indeed achieves better training effects than PINN when dealing with the more complex heterogeneous two-phase oil-water seepage process in the physical space in this example.

We compute the reference solution for this example using the upwind FDM with the same grid points and time steps as Example 3. Fig. 20 compares the predicted oil phase pressure distributions and the absolute error distributions of the pressure predictions by the three QCPINNs and the traditional PINN with the reference solution at $t$=0.82 (according to unit conversion, one unit time in this example corresponds to 100000/864≈115.7407 days). From the perspective of absolute error distribution, the absolute error of the pressure prediction by PINN is significantly higher overall than that of the three QCPINN models, with its error peak exceeding 0.117 MPa, while the absolute errors of the pressure predictions by the three QCPINN models are all less than 0.039 MPa.

Fig. 21 compares the predicted water saturation distributions and the absolute error distributions of the saturation predictions by the three QCPINNs and PINN with the reference solution at $t$=0.82. In terms of saturation distribution, the results of the QCPINN models are obviously more consistent with the reference solution than those of the PINN model, and the numerical diffusion of the QCPINN results at the water flooding front is significantly smaller than that of PINN. The error distribution diagrams also clearly reflect that the absolute error values of PINN are significantly higher, which demonstrates the superior performance of QCPINN.

Fig. 22 shows the evolution law of the L2 error of the oil phase pressure distribution and water saturation distribution predicted by the three QCPINN models and PINN within the time range of 0-1. It can be seen that the error of PINN remains at the highest level throughout the time range of 0-1, while the L2 errors of the three QCPINN models are significantly lower than that of PINN all the time. Among them, the average error level of the Cross-mesh topology is the lowest among the three quantum circuit topologies. In terms of pressure distribution, the L2 error of PINN is nearly 20 times that of QCPINN at maximum; in terms of saturation distribution, the L2 error of PINN is nearly 5 times that of QCPINN at maximum.

In general, for the two-phase flow problem with strongly heterogeneous permeability distribution and full pressure-saturation coupling in this example, the three QCPINN models all exhibit better training convergence and prediction accuracy compared with PINN, which fully demonstrates the important significance of the QCPINN studied in this paper to replace PINN and introduce quantum computing into the field of classical models and reservoir engineering.

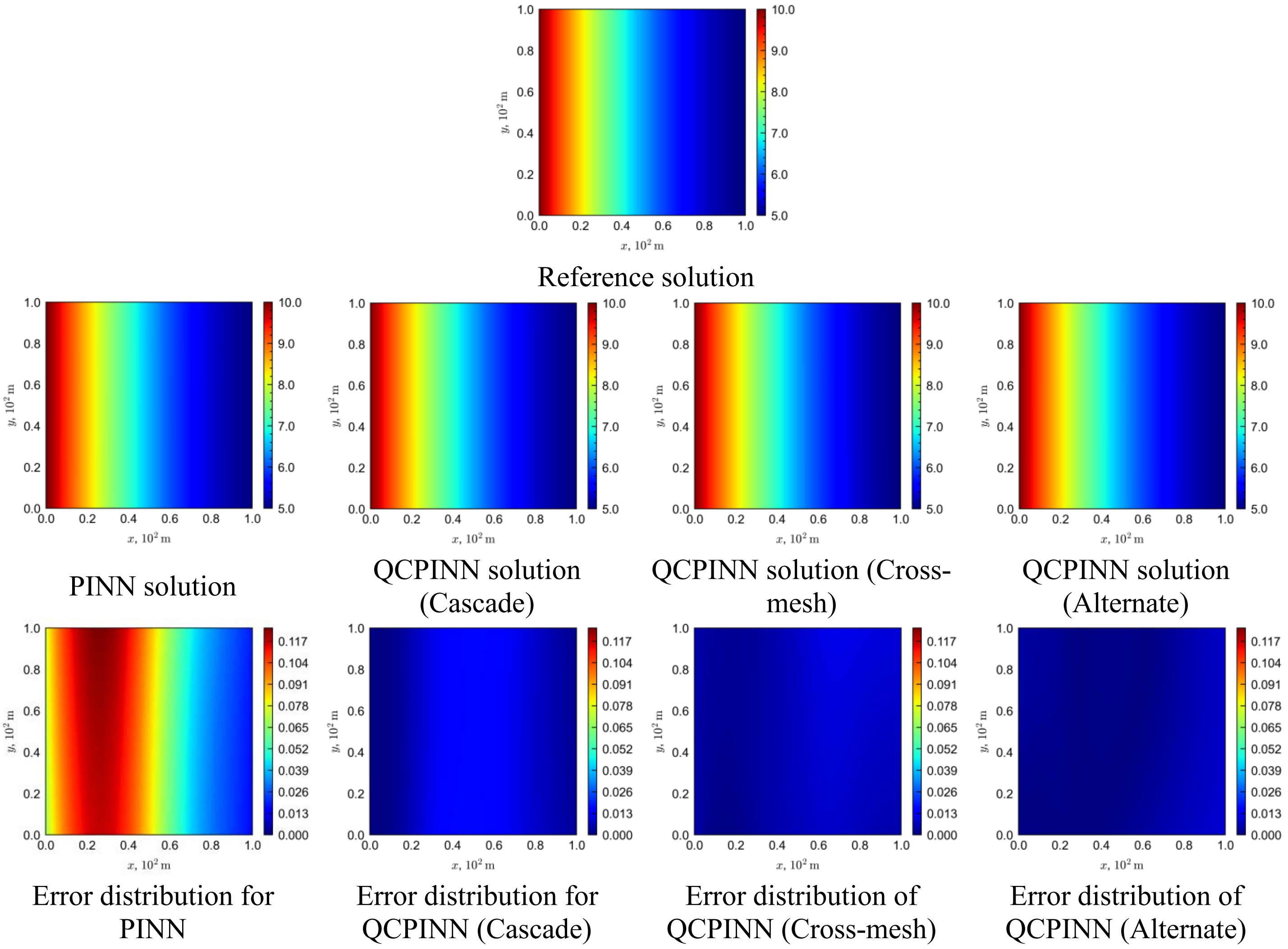

Fig. 20 Comparison of pressure distributions and absolute error distributions of QCPINN predictions using three topologies and PINN prediction at $t$=0.82 in Example 4.

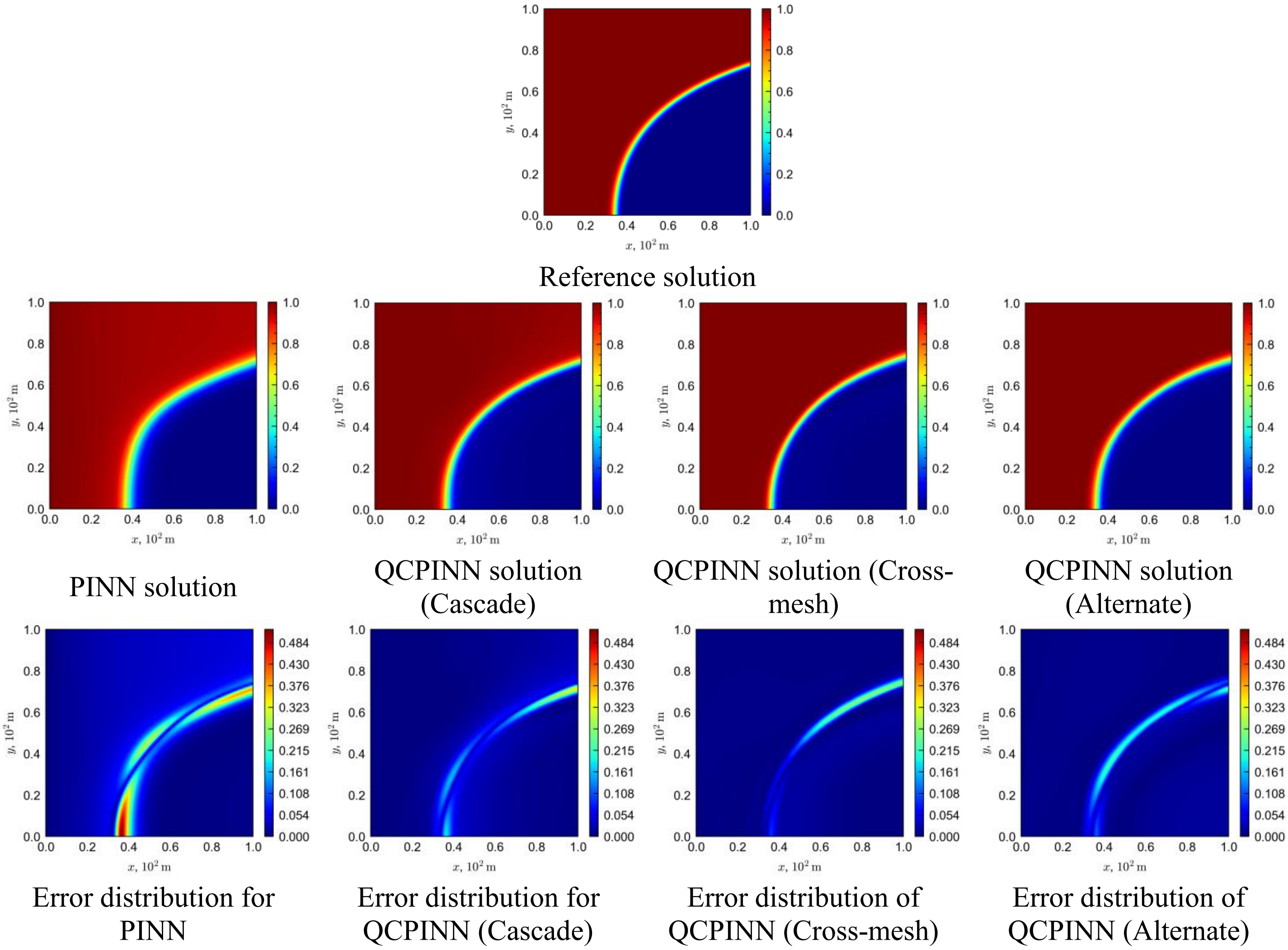

Fig. 21 Comparison of water saturation distributions and absolute error distributions of QCPINN predictions using three topologies and PINN prediction at $t$=0.82 in Example 4.

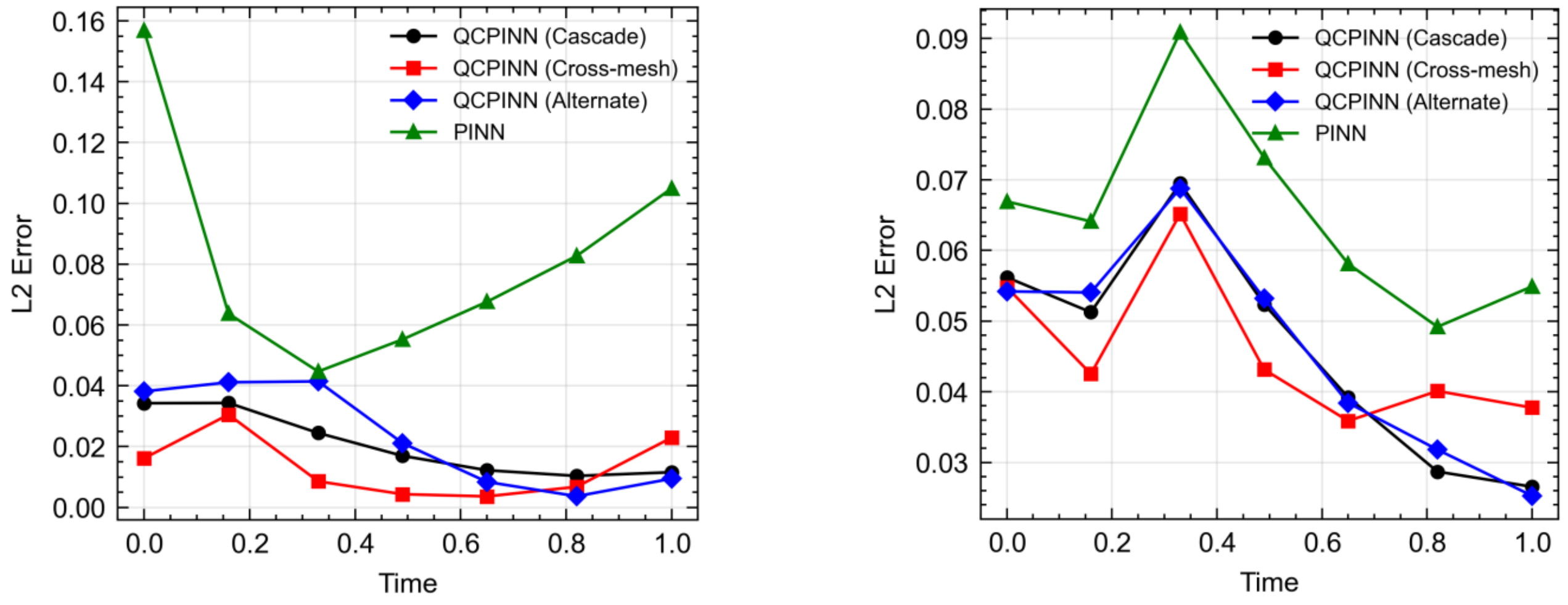

(a) Pressure

(b) Water saturation

Fig. 22 Comparison of L2 errors of pressure and water saturation predictions: QCPINN with three topologies vs PINN in Example 4.

### 4.5 Discussion on the Relationship Between Quantum Circuit Topology and Physical Problem Characteristics

Divergent optimal quantum circuit topologies across four seepage PDEs may reflect the intrinsic link between circuit design and the mathematical-physical traits of governing equations. This exploratory work provides interpretative insights to guide future quantum-classical co-design for reservoir simulation, though it cannot draw definitive causal rules. Circuit topology likely dominates QCPINN training behavior such as convergence rate and loss fluctuation, and we set unified training epochs for fair comparison. Entanglement layout, parameters and circuit depth may jointly determine feature extraction capacity, which could explain why the three topologies exhibit distinct performance in resolving spatial correlations, transient variations and steep physical fronts, as observed in our numerical experiments.

The elliptic pressure diffusion equation in Eq. (2) features global long-range pressure responses triggered by local permeability heterogeneities. The Cross-mesh topology achieves the highest accuracy, followed by Alternate, while Cascade yields larger errors. It is possible that Cross-mesh's full entanglement effectively captures elliptic global coupling; Alternate may deliver stable yet incomplete global field characterization, and Cascade's cyclic entanglement might struggle to reflect far-field heterogeneity impacts.

The hyperbolic Buckley-Leverett equation in Eq. (3) contains moving shock discontinuities and nonlinear saturation relations. The Alternate topology balances convergence stability and front capture via staggered local entanglement and may avoid the front smearing issue seen with Cascade. Cross-mesh possesses strong expressive power yet tends to generate large late-stage transient errors, which could stem from full-connectivity-induced overfitting and optimization difficulties. This observation suggests staggered local entanglement may be more suitable for nonlinear hyperbolic systems with sharp discontinuities.

The convection-diffusion-adsorption equation in Eq. (6) couples advection, diffusion and adsorption retardation, and the Alternate topology performs best for this multi-physics system. Its staggered local entanglement may simultaneously capture local diffusion and streamwise long-range advection that align with transport-retardation mechanisms. Cross-mesh may converge unstably possibly due to excessive parameters and sensitive gradient optimization, and Cascade's cyclic entanglement might poorly resolve long-range advection and introduce obvious errors near concentration fronts.

The fully coupled two-phase seepage equation in Eq. (31) is the most complex case combining elliptic pressure diffusion, hyperbolic saturation transport, heterogeneity and two-field coupling. The Cross-mesh topology attains the lowest overall $L_2$ errors for both pressure and saturation across all timestamps, which may be attributed to its cross-field full entanglement that enables balanced prediction of the two physical fields. The Alternate topology can resolve sharp waterfronts but may yield slightly higher overall errors than Cross-mesh in this coupled scenario, while Cascade delivers acceptable pressure predictions yet tends to blur saturation fronts, potentially because its cyclic entanglement is only capable of characterizing single-field correlations.

## 5 Conclusions and Future work

This study pioneers the adaptation and application of Quantum-Classical Physics-Informed Neural Networks (QCPINNs) to reservoir numerical simulation, systematically solving four core seepage PDEs featuring reservoir heterogeneity, strong nonlinearity, convection-dominated transport and full pressure-saturation coupling. The key conclusions are summarized as follows.

The improved DV-Circuit QCPINN unites quantum advantages with physical constraints, adopting classical networks for feature preprocessing and decoding alongside a quantum core for high-dimensional mapping to resolve typical defects of classical PINNs and support accurate PDE solving. Numerical tests over four representative reservoir seepage models verify that QCPINNs can faithfully reproduce key flow behaviors, including pressure distributions in heterogeneous media, steep saturation fronts during waterflooding, adsorption-delayed solute migration in compositional flow, and coupled pressure-saturation evolution in heterogeneous two-phase flow with exponential permeability fields. Quantitative error evaluation demonstrates that all QCPINN variants deliver low prediction errors and consistently outperform classical

PINNs; for the fully coupled two-phase flow case specifically, QCPINNs reduce prediction errors by up to 20 times for pressure and 5 times for water saturation.

Further comparative analysis of three quantum circuit topologies reveals case-dependent performance and distinct applicable scenarios. The Cross-mesh topology attains the minimal overall $L_2$ error across all four test cases, showing great adaptability to heterogeneous single-phase flow and stable performance for multi-physics coupled pressure-saturation systems, which may be attributed to its fully connected entanglement layout. While the Alternate topology is optimal for convection-diffusion-adsorption compositional flow and nonlinear Buckley-Leverett equations thanks to its superior front-capturing ability, it produces marginally larger errors than Cross-mesh within fully coupled two-phase systems. The Cascade topology delivers moderate prediction accuracy across all test cases, as its cyclic entanglement structure limits its capacity to characterize long-range advection and sharp transient concentration fronts.

This work extends QCPINN from simple fundamental PDEs to mainstream industrial reservoir seepage problems, establishing a novel technical framework for efficient reservoir simulators and data-driven surrogate models. The proposed QCPINN architecture is compatible with existing PINN optimization strategies, laying a solid foundation for follow-up research targeting practical complex reservoirs with intense heterogeneity, intricate fracture networks, multiphase-multicomponent coupled flow, transient displacement and complex well arrangements, which further broadens its industrial applicability.

For future work, we will first optimize model scalability and convergence by designing anti-barren-plateau quantum circuits, robust initialization strategies and advanced optimizers to escape local minima, alongside systematic scaling experiments with more qubits and circuit layers to fully unlock the universal approximation potential of QCPINNs. Beyond structural and training optimization, we plan to integrate mature quantum noise reduction strategies into the QCPINN framework, including shot-averaging optimization and variance-reduction measurement schemes to suppress sampling noise during measurement, quantum error mitigation tools like zero-noise extrapolation to counteract hardware decoherence and sampling bias, and customized noise-aware loss functions for finite-shot measurements that enable stable learning of physical constraints amid noisy measurement results.

## 6 Acknowledgements

Prof. Xiang Rao gratefully acknowledges the financial support from the National Science and Technology Major Project (Grant No. 2025ZD1401106), the General Program of Natural Science Foundation of Hubei Province (No. 2026AFB760), the General Program of the National Natural Science Foundation of China (Grant No. 52574028), and the "Science and Technology Innovation Team" Program of the Xinjiang Uygur Autonomous Region (Grant No. 2024TSYCTD0018).